\PassOptionsToPackage{table}{xcolor}
\PassOptionsToPackage{hyphens}{url}
\documentclass{article}
\usepackage{textcomp}

\usepackage{PRIMEarxiv}
\usepackage{tikz}
\usepackage{lmodern}
\usetikzlibrary{shapes,arrows}
\usepackage{multirow}
\usepackage{longtable}
\usepackage{booktabs}
\usepackage{adjustbox}
\usepackage{tabularx}
\usepackage{adjustbox}
\usepackage{balance}
\usepackage{colortbl}
\usepackage{xcolor}
\usepackage{lscape}
\usepackage[table]{xcolor}
\usepackage{graphicx}
\usepackage{forest}
\usepackage{algpseudocode}
\usepackage{subcaption}
\usepackage{hyperref}
\usepackage{xurl}
\usepackage[hyphens]{url}
\usepackage{color, colortbl}
\usepackage[most]{tcolorbox}
\usepackage{longtable}
\usepackage{float}
\usepackage{lscape}
\usepackage{booktabs}
\usepackage{array}
\usepackage{multirow}
\usepackage{tcolorbox}
 \usepackage{amsmath}
\usepackage{booktabs}
\usepackage[most]{tcolorbox}
\newcounter{highlightbox}
\newtcolorbox[use counter=highlightbox]{highlightbox}[2][]{%
  colback=blue!4!white, 
  colframe=blue!70!black, 
  title=Box~\thehighlightbox: #2,
  fonttitle=\bfseries,
  #1
}

\usepackage{tocloft}
\usepackage{hyperref}
\hypersetup{
    colorlinks=true,
    linkcolor=blue,
    urlcolor=blue,
    pdfborder={0 0 0}
}
\usepackage{tabularx}
\usepackage{fancyhdr}  % For fancy headers and footers
\tcbset{
    myboxstyle/.style={
        enhanced,
        colback=gray!5!white,
        colframe=black!40,
        boxrule=0.4pt,
        arc=5pt,
        boxsep=5pt,
        left=8pt,
        right=8pt,
        top=6pt,
        bottom=6pt,
        sharp corners=south,
        fonttitle=\bfseries,
        before skip=10pt,
        after skip=10pt
    }
}
\definecolor{pastelBlue}{HTML}{D0E3F7}
\definecolor{pastelGreen}{HTML}{D0F0D8}
\definecolor{pastelOrange}{HTML}{FFE5C4}
\definecolor{pastelPurple}{HTML}{E0D4F7}

\newtcolorbox{promptbox}[2][]{
  enhanced,
  breakable,
  colback=#2,               % pastel background
  colframe=#2!60!black,     % slightly darker frame
  colbacktitle=#2!30!white, % title background
  coltitle=black,
  fonttitle=\small\bfseries,
  title={#1},
  arc=1mm,                  % slightly rounded corners
  boxrule=0.7pt,            % border thickness
  boxsep=1pt,               % inner separation between text and border
  left=3pt,                 % left padding
  right=3pt,                % right padding
  top=3pt,                  % top padding
  bottom=3pt,               % bottom padding
  before skip=4pt,          % space above each box
  after skip=4pt,           % space below each box
  width=\textwidth          % make each box span the full text width
}	
\definecolor{Gray}{gray}{0.9}

\usepackage[most]{tcolorbox}      % loads tcolorbox with ‘enhanced’
\definecolor{mybluebg}{HTML}{f3f8fb}    % light gray background
\definecolor{myclueborder}{HTML}{6faad0} % light gray border
\definecolor{mybluetitle}{HTML}{bcd7e9}  % title background
\definecolor{myshadow}{HTML}{d8d8d8}

\usepackage{tikz}
\usetikzlibrary{positioning}

\usepackage[T1]{fontenc}
\usepackage{xparse}
\usepackage{enumitem}
\setlist[description]{
  font={\sffamily\bfseries},
  labelsep=0pt,
  labelwidth=\transcriptlen,
  leftmargin=\transcriptlen,
}

\newlength{\transcriptlen}

\NewDocumentCommand {\setspeaker} { mo } {%
  \IfNoValueTF{#2}
  {\expandafter\newcommand\csname#1\endcsname{\item[#1:]}}%
  {\expandafter\newcommand\csname#1\endcsname{\item[#2:]}}%
  \IfNoValueTF{#2}
  {\settowidth{\transcriptlen}{#1}}%
  {\settowidth{\transcriptlen}{#2}}%
}
\renewcommand{\arraystretch}{1.2}
\setspeaker{Q}
\setspeaker{A}[A]

\title{Thinking Beyond Tokens: From Brain-Inspired Intelligence to Cognitive Foundations for Artificial General Intelligence and its Societal Impact}
\author{
Rizwan Qureshi\textsuperscript{1}\thanks{Equal Contribution}, 
Ranjan Sapkota\textsuperscript{2}\footnotemark[1], 
Abbas Shah\textsuperscript{3}\footnotemark[1], 
Amgad Muneer\textsuperscript{4}\footnotemark[1],\\
Anas Zafar\textsuperscript{4},
Ashmal Vayani\textsuperscript{1},
Maged Shoman\textsuperscript{5},
Abdelrahman B. M. Eldaly\textsuperscript{6},
Kai Zhang\textsuperscript{4}, \\
Ferhat Sadak\textsuperscript{7},
Shaina Raza\textsuperscript{8}\thanks{Corresponding author: shaina.raza@torontomu.ca}, 
Xinqi Fan\textsuperscript{9},
Ravid Shwartz-Ziv\textsuperscript{10},
Hong Yan\textsuperscript{6},
Vinjia Jain\textsuperscript{11},\\
Aman Chadha\textsuperscript{12},
Manoj Karkee\textsuperscript{2},
Jia Wu\textsuperscript{4},
and 
Seyedali Mirjalili\textsuperscript{14,15}
\thanks{ \footnotesize
\textsuperscript{1} Center for research in Computer Vision, University of Central Florida, Orlando, FL, USA. 
\textsuperscript{2} Cornell University, Department of Biological and Environmental Engineering, Ithaca, NY 14853, USA
\textsuperscript{3} Department of Electronics Engineering, Mehran University of Engineering \& Technology, Jamshoro, Sindh, Pakistan.
\textsuperscript{4}Department of Imaging Physics, The University of Texas MD Anderson Cancer Center, Houston, TX, USA.
\textsuperscript{5} Intelligent Transportation Systems, University of Tennessee, Oakridge, TN, USA.
\textsuperscript{6} Department of Electrical Engineering, City University of Hong Kong, SAR China.
\textsuperscript{7} Department of Mechanical Engineering, Bartin University, Bartin Turkey.
\textsuperscript{8} Vector Institute, Toronto Canada.
\textsuperscript{9} Manchester Metropolitan University, Manchester, UK.
\textsuperscript{10} Center for Data Science, New York University, NYU, NY, USA.
\textsuperscript{11} Meta Research (Work done outside Meta).
\textsuperscript{12} Amazon Research (Work done outside Amazon).
\textsuperscript{14} Centre for Artificial Intelligence Research and Optimization, Torrens University Australia, Fortitude Valley,
Brisbane, QLD 4006, Australia,
\textsuperscript{15} University Research and Innovation Center, Obuda University, 1034 Budapest, Hungary}
}

\begin{document}
\maketitle
\vspace{-0.5em}

\begin{abstract}
\footnotesize
\vspace{-0.5em}

Can machines truly think, reason and act in domains like humans? This enduring question continues to shape the pursuit of Artificial General Intelligence (AGI). Despite the growing capabilities of models such as GPT-4.5, DeepSeek, Claude 3.5 Sonnet, Phi4, and Grok 3, which exhibit multimodal fluency and partial reasoning, these systems remain fundamentally limited by their reliance on token-level prediction and lack grounded agency. This paper offers a cross-disciplinary synthesis of AGI development, spanning artificial intelligence, cognitive neuroscience, psychology, generative models, and agent-based systems. We analyze the architectural and cognitive foundations of general intelligence, highlighting the role of modular reasoning, persistent memory, and multi-agent coordination. In particular, we emphasize the rise of Agentic RAG frameworks that combine retrieval, planning, and dynamic tool use to enable more adaptive behavior. We discuss generalization strategies, including information compression, test-time adaptation, and training-free methods, as critical pathways toward flexible, domain-agnostic intelligence. Vision-Language Models (VLMs) are reexamined not just as perception modules but as evolving interfaces for embodied understanding and collaborative task completion. We also argue that true intelligence arises not from scale alone but from the integration of memory and reasoning: an orchestration of modular, interactive, and self-improving components where compression enables adaptive behavior. Drawing on advances in neurosymbolic systems, reinforcement learning, and cognitive scaffolding, we explore how recent architectures begin to bridge the gap between statistical learning and goal-directed cognition. Finally, we identify key scientific, technical, and ethical challenges on the path to AGI, advocating for systems that are not only intelligent but also transparent, value-aligned, and socially grounded. We anticipate that this paper will serve as a foundational reference for researchers building the next generation of general-purpose human-level machine intelligence. \\The project can be accessed at \href{https://github.com/anas-zafar/agi-cognitive-foundations}{this GitHub repository}.

\end{abstract}

\vspace{-0.5em}


% -------- Inside the document --------

\onecolumn
\begin{center}
    {\Large \textbf{Table of Contents}}
     \vspace{-0.4em}
\end{center}
\tiny
\tableofcontents
\clearpage

\twocolumn
\normalsize

\vspace{0.5em}

\section{Introduction}

\emph{Can machines truly think?} Over seven decades ago, Alan Turing famously posed this foundational question at the dawn of computing. It remains central to the field of Artificial General Intelligence (AGI), which seeks to replicate the full breadth of human cognitive abilities in computational form~\cite{turing2009computing}. Yet, despite decades of progress, the term “thinking”~\cite{bruner2017study} itself is often invoked without sufficient precision~\cite{pfeifer2001understanding}. To meaningfully address this question, we must first define what we mean by thinking and related concepts, such as consciousness, intelligence, and generalization:
\begin{itemize}
    \item \textbf{Thinking}: The Manipulation of internal representations to solve problems, reason about the world, and generate novel ideas~\cite{bruner2017study}.
    \item \textbf{Consciousness}: The subjective capacity for awareness and self-reflection~\cite{tononi2015consciousness}.
    \item \textbf{Intelligence}: The capacity to acquire, apply, and adapt knowledge across tasks and environments~\cite{pfeifer2001understanding}.
    \item \textbf{AGI}: Systems capable of broad, human-level reasoning and learning across domains, without the need for task-specific retraining~\cite{khan2024agi}.
\end{itemize}

While leading-edge AI models such as GPT-4~\cite{firat2023if}, DeepSeek~\cite{liu2024deepseek}, and Grok~\cite{jegham2025visual} have demonstrated impressive performance across a diverse array of specialized tasks, their underlying architecture remains fundamentally limited by token-level prediction. Although this paradigm excels at surface-level pattern recognition, it lacks grounding in physical embodiment, higher-order reasoning, and reflective self-awareness, which are the core attributes of general intelligence~\cite{qi2024next}. Furthermore, these models do not exhibit consciousness or an embodied understanding of their environment, limiting their ability to generalize and adapt effectively to novel,  open and real-world scenarios~\cite{li2024embodied}.

% \begin{tcolorbox}[colback=blue!5!white, colframe=blue!75!black, top=2pt, bottom=2pt, title=Why token-level next-word prediction alone is insufficient for AGI:]  
% Token-level next-word predictions lack grounding in physical reality, fail to support causal reasoning, abstraction, or self-reflection, and cannot model multi-modal, goal-directed behavior essential to general intelligence.
% \end{tcolorbox}

\begin{tcolorbox}[colback=mybluebg, colframe=myclueborder, colbacktitle=mybluetitle, coltitle=black, fonttitle=\bfseries, top=2pt, bottom=2pt, title=Why Token-level Next-word Prediction Alone is Insufficient for AGI?]  
Next-token prediction models capture surface linguistic patterns but fail to support complex mental representations grounded in the physical world. Lacking embodiment, causality, and self-reflection, they struggle with abstraction and goal-directed behavior—core requirements for AGI.
\end{tcolorbox}

Post-training strategies~\cite{kumar2025llm} such as instruction tuning~\cite{liu2023visual} and Reinforcement Learning with Human Feedback (RLHF)~\cite{kaufmann2023survey} improve alignment and usability,
but operate within the same autoregressive framework. They introduce behavioral refinements,
not architectural changes~\cite{kaufmann2023survey}. Consequently, despite post-training advances, these models remain limited in their capacity to generalize in the open-ended, compositional manner characteristic of AGI~\cite{qi2024next}.

% \begin{tcolorbox}[colback=blue!5!white, colframe=blue!75!black, top=2pt, bottom=2pt, title=Why Post-Training and Alignment Can’t Bridge the Gap to AGI]  
\begin{tcolorbox}[colback=mybluebg, colframe=myclueborder, colbacktitle=mybluetitle, coltitle=black, fonttitle=\bfseries, top=2pt, bottom=2pt, title=Why Post-Training and Alignment Can’t Bridge the Gap to AGI?]  
Post-training methods, such as, Instruction tuning and RLHF transformed base models like GPT into more usable agents like ChatGPT. However, these alignment methods operate on top of token-level prediction and cannot endow models with core AGI traits—such as abstraction, grounded reasoning, or environmental awareness.

\end{tcolorbox}

Although model scaling can approximate complex representations and produce emergent behaviors, it lacks inductive biases for structured reasoning, fails to support persistent memory, and cannot generate self-models or agency. These limitations are architectural, not parametric—hence, scaling alone yields diminishing returns and cannot achieve AGI~\cite{shanmugam2022learning, shang2024ai}.

% \begin{tcolorbox}[colback=blue!5!white, colframe=blue!75!black, top=2pt, bottom=2pt, title=Why Further Scaling Will Not Lead to AGI]  

\begin{tcolorbox}[colback=mybluebg, colframe=myclueborder, colbacktitle=mybluetitle, coltitle=black, fonttitle=\bfseries, top=2pt, bottom=2pt, title=Why Further Scaling Will Not Lead to AGI?]  
While scaling improves fluency and performance on many tasks, it cannot resolve core limitations of current LLMs. These models still lack grounded understanding, causal reasoning, memory, and goal-directed behavior. 
\end{tcolorbox}

Besides next-token prediction, trajectory modeling frameworks (e.g. Algorithm 1), such as, The~\textit{Decision Transformer} reframe reinforcement learning as conditional sequence modeling, enabling policy generation via trajectory-level representations optimized for long-term return~\cite{chen2021decision}. Complementarily, \textit{self-prompting} mechanisms introduce latent planning loops~\cite{xie2025latent}, wherein models generate internal scaffolds to structure multi-step reasoning~\cite{yao2023react}. \textit{DeepSeek-V2}, a 236B-parameter Mixture-of-Experts model with a 128K-token context, exemplifies this paradigm by integrating trajectory modeling with reinforcement fine-tuning to improve coherence and planning across extended tasks~\cite{lu2024deepseek}. Collectively, these approaches advance beyond token-level generation by embedding structured, goal-conditioned reasoning within the model architecture~\cite{yao2023react}.

\begin{tcolorbox}[enhanced, sharp corners, colback=gray!5!white, colframe=blue!80!black,
title=Algorithm 1: Trajectory-Based Planning via Decision Transformers, fonttitle=\bfseries, boxrule=0.5pt]
\textbf{Input:} Goal \( G \), history \( H \), reward function \( R \) \\
\textbf{Output:} Action sequence \( A = \{a_1, a_2, \ldots, a_T\} \)

\begin{enumerate}
    \item Encode history and desired return into trajectory-level input
    \item Use Decision Transformer to predict next actions conditioned on future reward
    \item Iteratively update sequence based on observed outcomes
    \item Integrate reward-to-go and attention over past states for long-horizon reasoning
    \item Output final plan \( A \)
\end{enumerate}
\end{tcolorbox}

% \begin{tcolorbox}[colback=blue!5!white, colframe=blue!75!black, top=2pt, bottom=2pt, title= Prompting-Based Scaffolds: Chain of Thought, Tree of Thoughts Prompting methods]  

\begin{tcolorbox}[enhanced, sharp corners, colback=gray!5!white, colframe=blue!80!black,
title=Algorithm 2: Prompt-Based Agentic Reasoning (CoT/ToT/ReAct), fonttitle=\bfseries, boxrule=0.5pt]
\textbf{Input:} Task description \( T \), retrieved context \( C \), agent memory \( M \) \\
\textbf{Output:} Solution \( S \) with intermediate reasoning steps

\begin{enumerate}
    \item Decompose task \( T \) into subproblems using Chain-of-Thought (CoT)
    \item Explore multiple reasoning paths via Tree-of-Thoughts (ToT)
    \item Interleave reasoning with tool/environment actions (ReAct)
    \item Score and revise trajectories based on feedback and self-evaluation
    \item Return final solution \( S \) and rationale trace
\end{enumerate}
\end{tcolorbox}

\textit{Chain-of-Thought} prompting further improves reasoning by decomposing tasks into interpretable sub-steps, enhancing performance on arithmetic, commonsense, and symbolic challenges~\cite{wei2022chain}. Extending this, the \textit{Tree-of-Thoughts (ToT)} framework enables large language models (LLMs) to explore and evaluate multiple reasoning paths via lookahead, backtracking, and self-evaluation, yielding significant gains in tasks requiring strategic planning~\cite{yao2023tree}. For instance, applying ToT to GPT-4 increased its success rate on a combinatorial puzzle from 4\% (CoT) to 74\%\cite{yao2023tree}. \textit{ReAct} further augments this space by interleaving reasoning with environment-aware actions, allowing models to iteratively gather information, revise plans, and improve factual accuracy~\cite{yao2023react}. These complementary methods collectively form the foundation of prompt-based agentic reasoning, enabling both structured internal deliberation and dynamic external interaction. A generalized overview of this unified reasoning process is presented in Algorithm 2.

% \textbf{note: Should we also have a note about test-time adaptation and scaling}
% This could be treated as part of the model scaling where we scaling at the test time, but there is no knowledge improvment

As AI systems increasingly influence healthcare, education, governance, and the labor market, their integration into society must be guided by ethical, inclusive, and equitable principles~\cite{hadi2023large}. Democratizing AI means equitably distributing access, participation, and benefits across regions, communities, and socioeconomic groups—narrowing existing disparities rather than reinforcing them~\cite{moon2023searching}.

% \begin{tcolorbox}[colback=blue!5!white, colframe=green!45!black, top=2pt, bottom=2pt, title=AI Integration and the Need for Democratization]

\begin{tcolorbox}[colback=mybluebg, colframe=myclueborder, colbacktitle=mybluetitle, coltitle=black, fonttitle=\bfseries, top=2pt, bottom=2pt, title=AI Integration and the Need for Democratization]  
Without inclusive development, AI may amplify existing inequalities and silence underrepresented voices. Trustworthy, transparent, and socially aligned systems are not optional; they are a societal necessity.
\end{tcolorbox}
%This paper provides a comprehensive synthesis of the limitations and future directions for AGI. We critically examine the limitations of current token-level prediction models, which excel at surface-level pattern recognition but lack grounded reasoning, planning, and adaptive capabilities essential for general intelligence. To address these gaps, we propose a framework grounded in cognitive science, neuroscience, and machine learning that prioritises modular cognition, grounded reasoning, memory-augmented systems, and multi-agent architectures. By evaluating enabling technologies such as neuro-symbolic systems, reinforcement learning, and multi-agent frameworks, we outline pathways to bridge the gap between narrow AI and human-like intelligence. Additionally, we address the societal and ethical challenges of AGI, offering actionable recommendations to ensure its safe, adaptive, and value-aligned growth.
Rodney Brooks in 2008 argued that intelligence emerges from physical embodiment rather than abstraction alone~\cite{brooks2008rodney}. Building on this and recent developments in AGI in cross-disciplinary domains~\cite{raman2025navigating}, we propose that AGI must arise through integrated perception, embodiment, and grounded reasoning, not scale alone. We synthesize decades of AGI research in machine learning, cognitive neuroscience, and computational theory, critically examining recent techniques such as Chain of Thought~\cite{wei2022chain}, Tree of Thoughts~\cite{yao2023tree}, ReAct~\cite{yao2023react}, and trajectory modeling~\cite{chen2021decision}. While these methods enhance structured reasoning, they remain transitional, lacking physical grounding, memory, and self-awareness—core to general intelligence~\cite{fengfar}.

To address these gaps, we explore neuro-symbolic systems, multi-agent coordination, and RLHF as building blocks of AGI. This review frames a roadmap toward systems that are cognitively grounded, modular, and value-aligned, centered on the question: What mechanisms are essential to move from prediction to general-purpose intelligence?

\subsection*{\textbf{Motivation}}
Artificial General Intelligence (AGI) aims to replicate the full spectrum of human cognition, including reasoning, learning, memory, perception, and adaptation in dynamic, open-ended environments~\cite{alam2022methodological}. It is widely regarded one of the most ambitious frontiers in science and technology~\cite{fengfar}, and interest in AGI continues to grow across academia and industry, with major contributions from OpenAI~\cite{achiam2023gpt}, Amazon~\cite{intelligence2025amazon}, Microsoft Research~\cite{abdin2024phi}, Google~\cite{team2023gemini}, and Meta~\cite{touvron2023llama}.

Although previous studies have explored AGI readiness~\cite{fengfar}, safety concerns~\cite{everitt2018agi}, applications in IoT~\cite{dou2023towards}, brain-inspired architectures~\cite{zhao2023brain}, and cognitive frameworks~\cite{leon2024review}, the fundamental challenge persists: how can we transition from statistical pattern recognition to machines capable of genuine reasoning and flexible generalization?

Recent models such as GPT-4, DeepSeek, and Grok demonstrate growing multimodal competence. However, they still lack core capabilities such as abstraction, grounded reasoning, and real-time adaptation, which are essential for building truly general intelligence.

%The structure of this paper is shown in Figure~\ref{fig:surveyoutline}, 

\paragraph{Key Contributions} To the best of our knowledge, this is the first review to evaluate AGI through three integrated lenses: computational architectures, cognitive neuroscience, and societal alignment. Specifically:
\begin{itemize}
    \item We introduce a unified framework that synthesizes insights from neuroscience, cognition, and AI to identify foundational principles for AGI system design.
    \item We critically analyze the limitations of current token-level models and post hoc alignment strategies, emphasizing the need for grounded, agentic, and memory-augmented architectures.
    \item We survey emergent AGI-enabling methods, including modular cognition, world modeling, neuro-symbolic reasoning, and biologically inspired architectures.
    \item We present a multidimensional roadmap for AGI development that incorporates logical reasoning, lifelong learning, embodiment, and ethical oversight.
    \item We map core human cognitive functions to computational analogues, offering actionable design insights for future AGI systems. A list of key acronyms used in this paper, are  defined in Appendix Table~\ref{tab:glossary}.
\end{itemize}

\section{Historical Evolution of AI}

AI has evolved through several major paradigms: from symbolic rule-based systems~\cite{duch2004computational} to statistical learning models~\cite{marra2024statistical}, and more recently into the era of generative and agentic AI~\cite{campbell2023computer}. As shown in Figure~\ref{fig:intro1}, modern generative models~\cite{West2023} excel at capturing data distributions and generating fluent text~\cite{yu2022survey}, speech~\cite{zhang2023speechgpt}, images and videos~\cite{jordon2022synthetic}, and even executable code~\cite{qi2024next}. Yet, despite their breadth, these systems remain fundamentally constrained: they operate at the level of token prediction, lacking grounded semantics, causal reasoning, and long-term planning~\cite{tong2024can}.

\par The emergence of more autonomous and general-purpose systems such as DeepSeek~\cite{lu2024deepseek}, GPT-4~\cite{bubeck2023sparks}, OpenAI's o1~\cite{wang2024planning}, DeepResearch and xAI's Grok3~\cite{jegham2025visual} signals a potential shift beyond static pattern matching. These models demonstrate early signs of multi-modal integration, creative problem-solving, and self-directed planning, pointing toward the first glimpses of general intelligence in machines.

\par Bridging the divide between narrow pattern-based intelligence and human-like generality is a central challenge for AGI~\cite{zhao2023brain}. A confluence of enabling technologies is accelerating this transition from generative AI to systems capable of adaptive, grounded, and goal-directed behavior~\cite{kanbach2024genai}. One fundamental thread is \textit{ deep reinforcement learning (RL)}~\cite{shakya2023reinforcement}, which enables agents to learn through trial-and-error interaction with dynamic environments. Landmark achievements, such as AlphaGo~\cite{lapan2018deep} and AlphaFold2~\cite{jumper2021highly}, illustrate how reinforcement learning and attention mechanisms support long-horizon decision-making and structural prediction. These systems rely on stable optimization methods such as \textit{Proximal Policy Optimization (PPO)}~\cite{schulman2017proximal}, which balances exploration with policy stability in high-dimensional action spaces.

\par To further align model behavior with human values, recent work emphasizes preference-based fine-tuning methods such as \textit{Direct Preference Optimization (DPO)}~\cite{rafailov2023direct} and \textit{Group Relative Policy Optimization (GRPO)}~\cite{shinn2023reflexion}. These techniques circumvent the need for explicit reward modeling by directly optimizing for human-aligned outcomes based on comparative preference signals. In parallel, \textit{neuro-symbolic systems}~\cite{garnelo2019reconciling} integrate symbolic reasoning with deep learning (DL), allowing agents to manipulate abstract variables and compositional rules. Collectively, these systems provide a path toward explainable and generalizable cognition, critical for robust AGI.

\subsection{Overview of AGI} AGI represents a frontier in the evolution of computational systems, striving to develop machines that can perform any intellectual task that a human can, across various domains~\cite{obaid2023machine}. Unlike narrow AI \cite{page2018risks}, which is designed for specific tasks, often operating on limited token-level inputs, AGI aims for a comprehensive cognitive ability, simulating the breadth and depth of human intellect \cite{braga2017emperor, kriegeskorte2018cognitive}. This ambition poses profound implications for society, promising revolutionary advances in healthcare~\cite{alam2022methodological}, education~\cite{siemens2022human}, and beyond~\cite{khan2024agi}, while also introducing complex ethical and safety challenges \cite{sonko2024critical}. AGI research encompasses diverse approaches, including symbolic~\cite{sheth2024neurosymbolic}, emergentist~\cite{firat2023if}, hybrid~\cite{sun2013connectionist}, and universalist models~\cite{tang2023agibench}, each offering distinct pathways toward achieving versatile intelligence \cite{goertzel2014artificial}. The development of AGI involves integrating sophisticated algorithms that can learn, reason, and adapt in ways that mimic human cognitive processes, such as learning from limited data~\cite{gruetzemacher2019toward}, transferring knowledge across contexts, and abstract reasoning \cite{laird2010cognitive,  thawakar2024mobillama}. Despite its potential, the field grapples with significant hurdles such as ensuring safety, managing unforeseen consequences, and aligning AGI systems with human values \cite{mahler2022regulating, obaid2023machine}. Furthermore, measuring progress in AGI development remains contentious, with debates over the appropriateness of benchmarks like the Turing Test \cite{moor2003turing} or operational standards akin to human educational achievements \cite{danziger2022intelligence}. As we advance, the integration of interdisciplinary insights from cognitive science, ethics, and robust engineering is crucial to navigate the complexities of AGI and harness its potential responsibly. 

\subsection{Agentic AI}

Although LLMs excel at predicting text, they lack the perceptual grounding that underpins human cognition~\cite{goertzel2023generative}. Humans build world models by continually integrating sensory input, memory, and action, skills rooted through direct, embodied interaction (e.g., a child learns to catch a ball by moving in space)~\cite{siemens2022human}. LLMs, by contrast, are disembodied: they cannot perceive, act, or internalize causal dynamics, so they struggle with tasks that demand physical reasoning, commonsense inference, or real-time adaptation \cite{van2021human}.

To address these limitations, a parallel frontier has emerged in the form of agentic architecture systems designed to perform autonomous planning, memory management, and inter-agent coordination \cite{sapkota2025vibe, sapkota2025ai}. A notable example is the Natural Language-based Society of Mind (NLSOM) framework~\cite{zhuge2023mindstorms}, which proposes a modular system composed of multiple specialized agents that communicate using natural language. These neural societies reflect Minsky's original vision~\cite{minsky1986society} of the mind as a collection of loosely coupled agents, each responsible for distinct cognitive tasks. By distributing intelligence across a community of specialized modules, NLSOM and similar architectures mitigate the monolithic limitations of conventional LLMs. They enable cognitive functions such as modular reasoning, episodic memory retrieval, and collaborative problem-solving traits essential for developing general-purpose intelligence~\cite{acharya2025agentic}.

\par
\begin{figure*}[htbp]
\centering
\includegraphics[width=\linewidth]{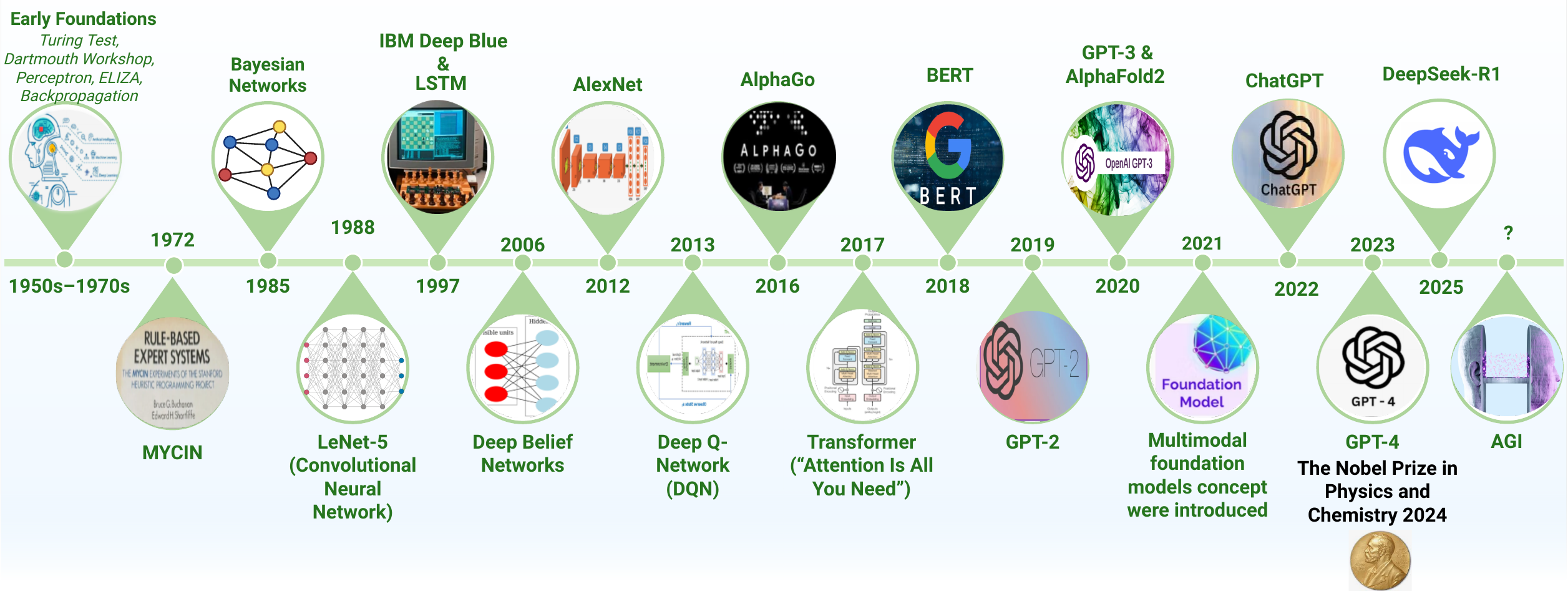}
\caption{A timeline of key milestones toward Artificial General Intelligence (AGI) from 1950 to 2025. The evolution spans symbolic systems (e.g., ELIZA), neural networks (e.g., LeNet-5, AlexNet), reinforcement learning (e.g., AlphaGo, DQN), foundation models (e.g., GPT-4, DeepSeek-R1), and \textbf{(Nobel Prize in Physics and Chemistry in 2024)}. This trajectory reflects a shift from static, rule-based methods to dynamic, multimodal, and increasingly general AI systems.}
\label{fig:intro1}
\end{figure*}

 These developments mark a transition from static, feedforward predictors to dynamic, interactive, and cognitively enriched AI systems~\cite{lampinen2021towards}. As depicted in Figure~\ref{fig:intro1}, AI has evolved from symbolic systems (e.g., Turing Test, ELIZA) to neural architectures (e.g., LeNet-5, Deep Belief Networks, AlexNet), then to reinforcement agents (e.g., DQN, AlphaGo), attention-based models (e.g., Transformer, BERT), and most recently, to foundation and emergent models such as GPT-4 and DeepSeek-R1.  A detailed chronology of modern AI and deep learning can be found in~\cite{schmidhuber2022annotated, haenlein2019brief}. 

\par Recent proposals such as S1 scaling\cite{liu2024deepseek} challenge the traditional focus on parameter count as the primary driver of AGI. Instead, they advocate scaling along cognitive axes—modularity, reasoning depth, self-prompting, and agentic coordination\cite{lu2024deepseek}. This structured approach marks a paradigm shift from undifferentiated statistical inference toward architecturally organized systems capable of flexible, interpretable reasoning~\cite{jegham2025visual}. Collectively, these trends signal a converging path toward open-ended, general-purpose machine intelligence.

  \begin{figure*}[htbp]
\centering
\includegraphics[width=0.98\linewidth]{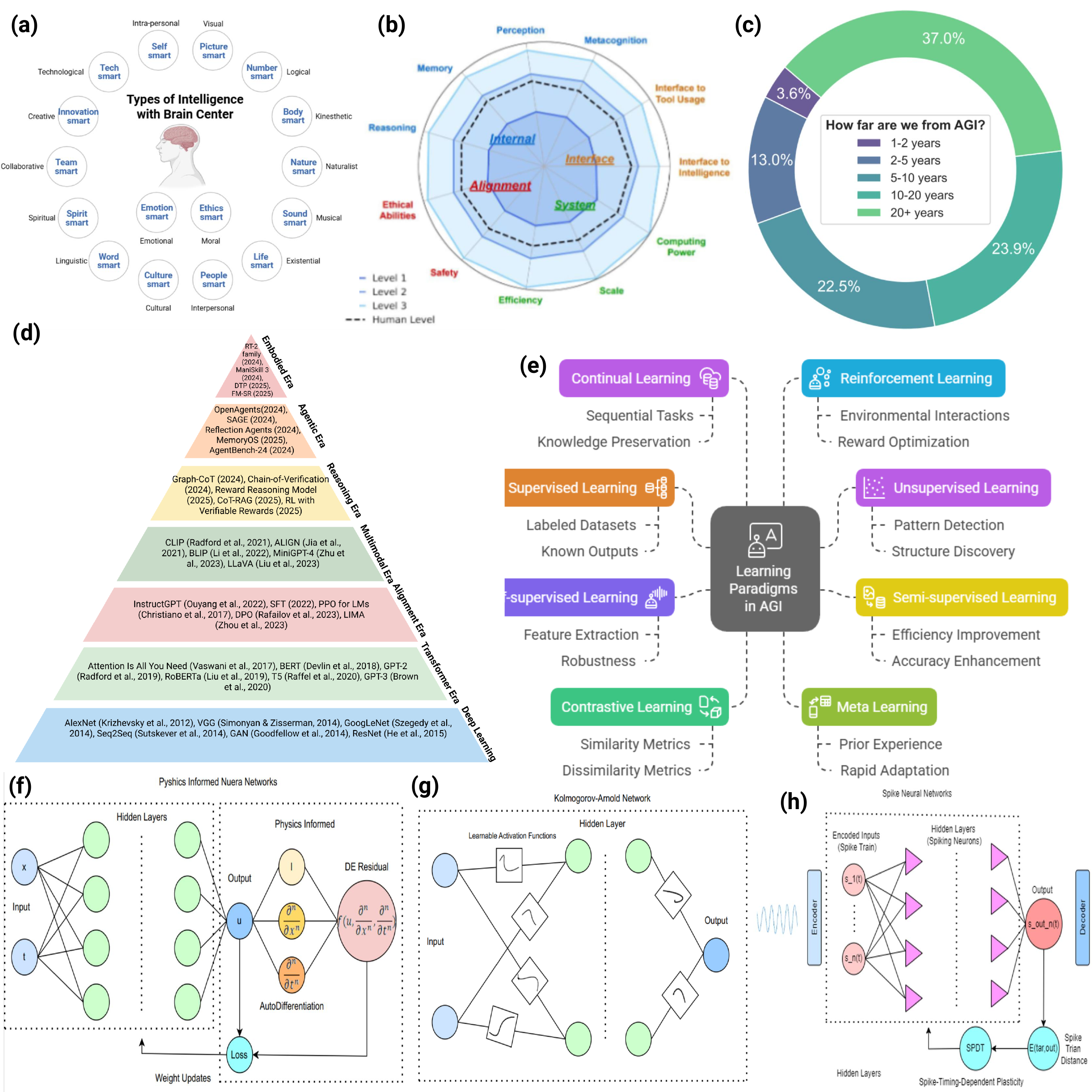}
\caption{An overview of foundational concepts, progress, and paradigms toward Artificial General Intelligence (AGI). (a) Multiple human intelligence types as conceptualized in brain-inspired AGI. (b) Radar chart representing the multidimensional alignment challenges in AGI including internal reasoning, external interface, system efficiency, and ethical safety. (c) Survey-based forecast of AGI timeline expectations adapted from ICLR 2024 survey~\cite{fengfar}. (d) Pyramid of Foundational AI Eras Leading to the Embodied Era. (e) Categorization of core learning paradigms in AGI, including supervised, unsupervised, self-supervised, and reinforcement learning, as well as emerging paradigms like continual, contrastive, semi-supervised, and meta learning. (f–h) Architectures representing (f) Physics-Informed Neural Networks (PINNs), (g) Kolmogorov–Arnold Networks (KANs), and (h) Spiking Neural Networks (SNNs) highlighting biological plausibility and adaptive computation in AGI development.}
\label{fig:intro}
\end{figure*}
\section{Understanding Intelligence - Logical Foundations of Intelligence} \label{Intelbasis}
Understanding the logical and cognitive foundations of intelligence is essential for developing robust AGI systems~\cite{zhu2023intelligent}. Intelligence covers diverse cognitive abilities, including perception, learning, memory, reasoning, and adaptability. Achieving AGI requires a comprehensive understanding of these cognitive processes and their neural bases~\cite{sun2006cognitive}.

\subsection{Brain Functionality}

The human brain, shown in Figure~\ref{fig:allnetworkq1}, is a highly intricate and partially understood organ that underlies core cognitive functions such as consciousness, adaptive intelligence, and goal-directed behavior~\cite{herculano2009human, chi2016neural}. Despite weighing only 1.3 to 1.5 kg, it accounts for nearly 20\% of the body's energy consumption, underscoring its metabolic and computational intensity~\cite{buxton2021thermodynamics, barros2018current}. Architecturally, the brain is organized into functionally specialized regions operating in tightly integrated hierarchies~\cite{chin2023beyond}. The neocortex a hallmark of mammalian evolution supports higher-order cognition and abstract reasoning, while subcortical structures regulate affective and autonomic functions~\cite{kanwisher2010functional}. Key components such as the hippocampus facilitate encoding of episodic memory (EM) and spatial navigation, whereas the occipital cortex governs visual processing and the motor cortex orchestrates voluntary movement~\cite{chin2023beyond}. These neurobiological insights offer design principles for AGI systems aiming to replicate cognitive flexibility, embodied intelligence, and adaptive decision-making.

The true computational power of the brain lies in its approximately 86 billion neurons, which create a dense network of about 150  trillion synaptic connections~\cite{azevedo2009equal, lent2025yes, buckner2013evolution}. This vast network enables both localized and extensive communications, positioning the brain as a complex, multi-scale network system. Synaptic activities, which include excitatory and inhibitory signals, maintain a critical balance essential for all cognitive functionalities \cite{navlakha2018network}. These synaptic interactions facilitate complex behaviors and thought processes, underscoring the importance of understanding these networks to replicate similar capabilities in AI systems \cite{liao2017small}. This neuro-computational foundation offers a road-map for developing AGI systems that aim to emulate human-like intelligence.
\subsubsection{Brain Functionalities and Their State of Research in AI}

\begin{figure*}
    \centering
    \includegraphics[width=0.98\linewidth]{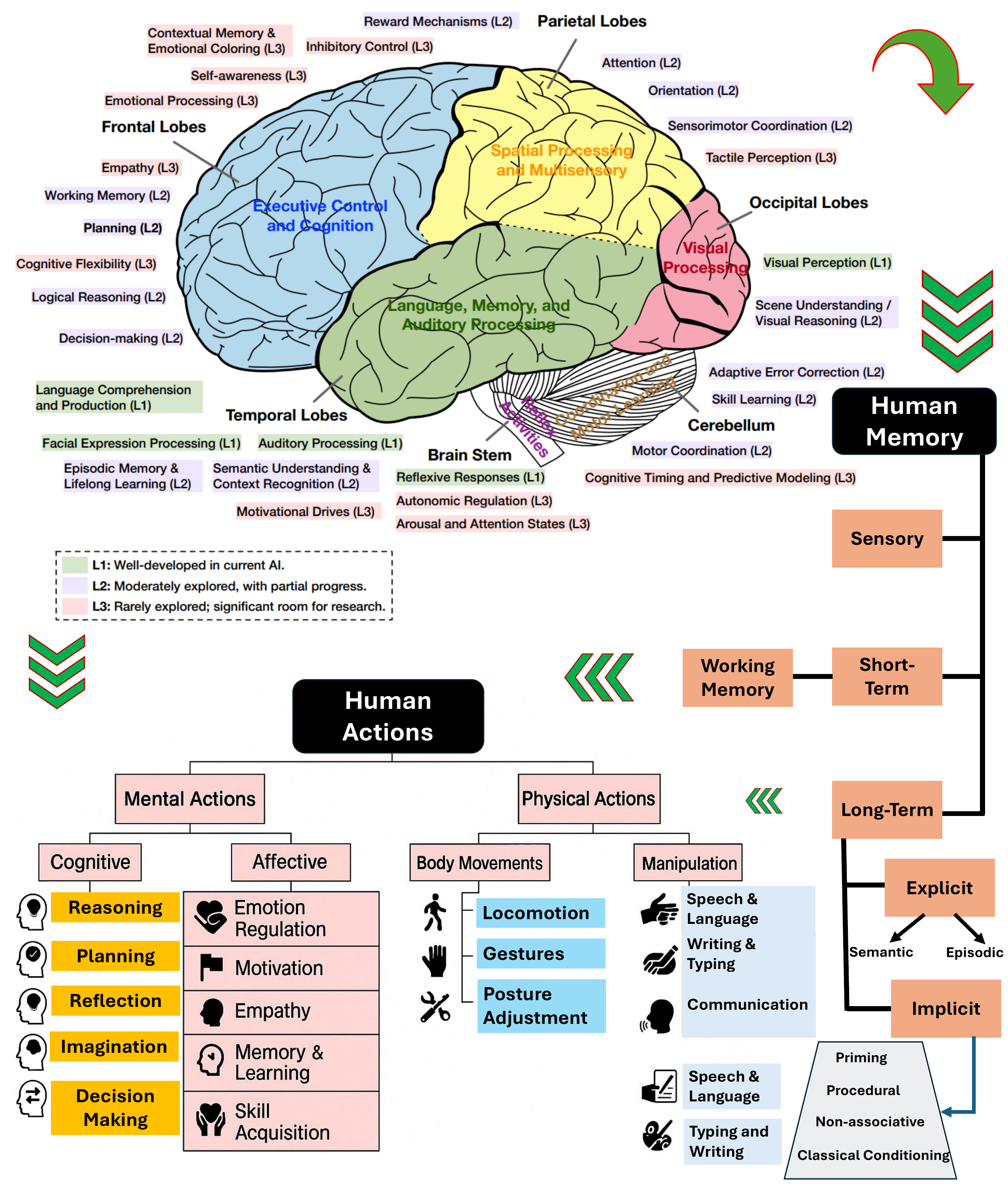}
    \caption{Illustrating the core cognitive pillars brain functions, memory hierarchies, action systems,    and world models that serve as foundational inspirations for AGI development. The upper section highlights key brain functionalities aligned with levels of AI research, identifying current achievements, gaps, and opportunities. It presents a hierarchical taxonomy of human memory, including sensory, short-term, working, and long-term types, further categorized into declarative and non-declarative forms. Additionally, it depicts human actions, spanning mental and physical dimensions crucial for cognition, planning, and goal-directed behavior. The brain diagram in this figure showing the functionalities of brain and their state of the research in AI is sourced from \cite{liu2025advances}.}
    \label{fig:allnetworkq1}
\end{figure*}

Figure~\ref{fig:allnetworkq1}a maps major brain regions to their AI counterparts, highlighting varying levels of research maturity: well-developed (L1), moderately explored (L2), and underexplored (L3). This comparison reveals both strengths and gaps in current AI research, offering a roadmap for advancing brain-inspired intelligence~\cite{liu2025advances}.
The frontal lobe governs high-level cognition such as planning and decision-making~\cite{frith1996role}, with AI showing strong performance in structured tasks (e.g., AlphaGo). Yet, traits like consciousness and cognitive flexibility remain underexplored (L3)~\cite{dolgikh2024self, na2024emergence}. In contrast, language and auditory functions mapped to L1 domains are well-modeled by LLMs, which approach human-level proficiency in language processing~\cite{liu2025advances, mahowald2024dissociating}.

Conversely, the cerebellum and limbic system govern fine motor skills and emotional processing, respectively~\cite{timmann2010human}. In AI, motor coordination is explored via robotics and meta-learning~\cite{miall2007cerebellum, sivalingam2024cerebellar}, yet achieving human-like dexterity and adaptability remains a challenge (L2–L3)~\cite{glickman2024human}. Emotional and motivational processes modeled by the limbic system are only superficially replicated in AI through reinforcement learning, highlighting a major gap in developing true emotional intelligence. (L3)~\cite{huang2024ai, ren2024brain}.

\begin{figure*}[htbp]
  \centering
  \includegraphics[width=0.88\textwidth]{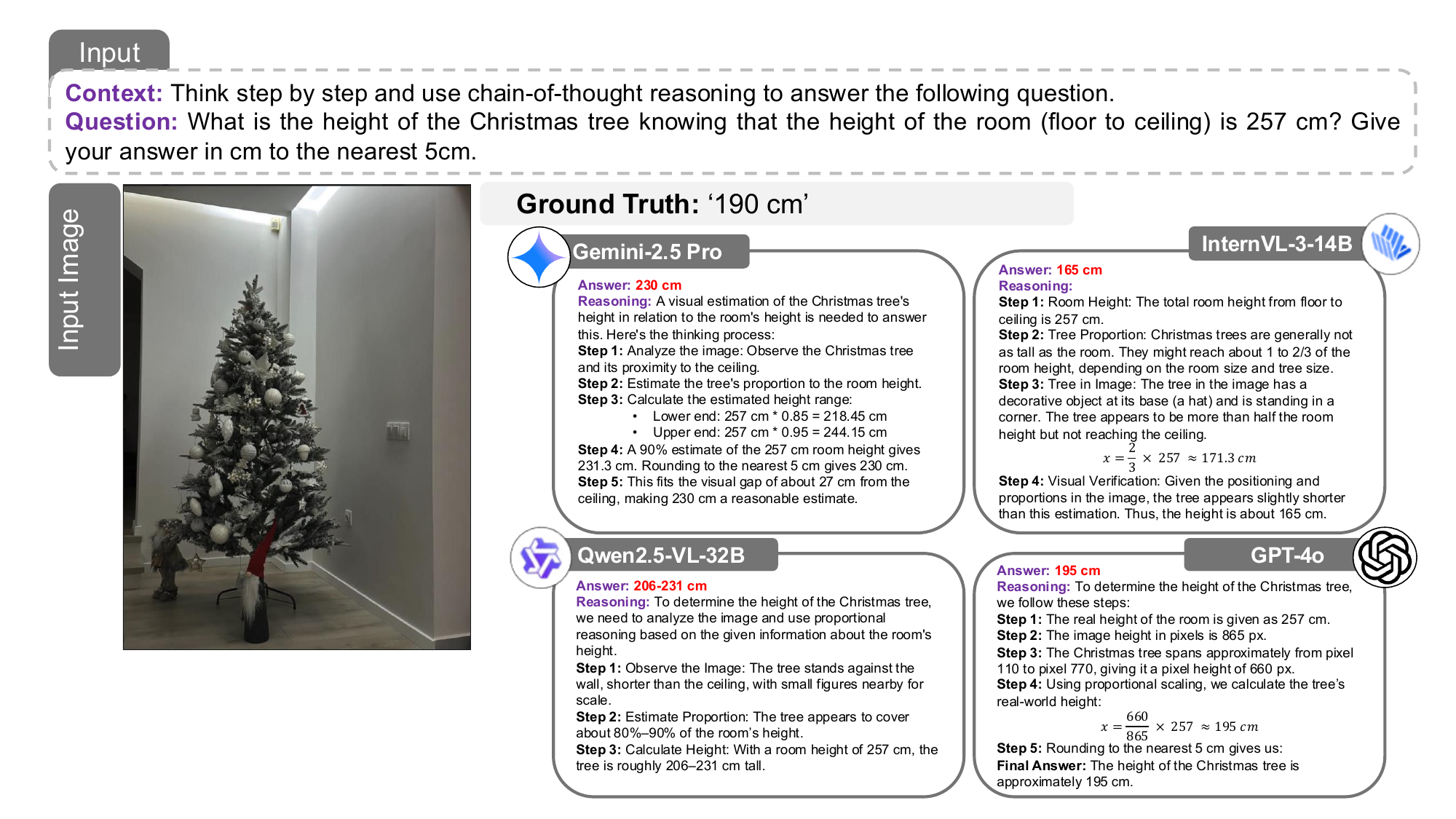}
  \caption{Illustration of the limitations of leading large multimodal models (LMMs) in performing accurate multi-step visual reasoning. Despite being prompted to follow a structured chain-of-thought, all models, ncluding Gemini-2.5 Pro, GPT-4o, Qwen-2.5-VL-32B, and InternVL-3-14B fail to estimate the Christmas tree height correctly based on the known room height of 257 cm. The ground truth of 190 cm highlights over- and under-estimations, exposing a persistent gap between visual perception, proportional reasoning, and precise numerical grounding in current LMMs.}
  \label{fig:Illustrationvisualization}
\end{figure*}

\subsubsection{Memory in Human and Artificial Intelligence}
Memory is a fundamental pillar of cognition in both humans and AI, enabling learning, adaptation, and problem-solving~\cite{gkintoni2025challenging}. In humans, it supports language acquisition, skill mastery, and social interaction core to self-awareness and decision-making~\cite{garfield2001social, greenspan2009first}. Likewise, in AI, memory facilitates intelligent behavior by supporting complex task execution, prediction, and adaptability~\cite{zheng2025machine}. This parallel underscores the value of biological memory insights in guiding the design of more advanced, memory-driven AI systems.
\par Figure~\ref{fig:allnetworkq1} presents a hierarchical taxonomy of human memory, outlining how sensory input transitions into short-term and long-term memory through encoding, consolidation, and retrieval~\cite{liu2025advances}. This framework offers a blueprint for AI memory systems, which have evolved from static data stores~\cite{molas2021advances, he2024human} to dynamic architectures that more closely mimic the flexibility and contextual awareness of human cognition.

Despite recent progress, AI memory systems still fall short of the contextual richness and adaptability of human memory~\cite{tariq2023technological}. Unlike humans, who integrate memory with perception, reasoning, and emotion~\cite{grossberg2020path}, AI typically relies on fixed algorithms and parameters. Achieving AGI will require memory systems that not only store information but also contextualize and conceptualize it akin to human cognition~\cite{li2025should}. Drawing from neuroscience and cognitive psychology such as the models in Figure~\ref{fig:allnetworkq1} offers a roadmap for building AI that learns from experience, adapts to new situations, and supports emotionally informed, lifelong learning~\cite{liu2025advances}.

\subsubsection{Human Action System: Mental and Physical Foundations for AGI}

The human action system comprising both mental and physical actions is central to intelligent behavior~\cite{vernon2007survey, wang2024parallel}. Mental actions include reasoning, planning, and memory recall, while physical actions encompass movement, communication, and interaction~\cite{liu2025advances} (Figure~\ref{fig:allnetworkq1}). Mental actions guide internal decision-making and simulate outcomes~\cite{andersen2009intention, crosato2024social}, whereas physical actions execute intentions and adapt behavior through real-world feedback~\cite{luck2000applying, van2021human}. This bidirectional loop between cognition and action provides a foundational model for AGI systems aiming to integrate perception, planning, and adaptive execution.

In AI agents, action systems are designed to emulate this cognitive loop \cite{colombo2021adaptive}. Language-based agents (e.g., using LLMs) simulate mental actions like reasoning and planning \cite{cheng2024exploring}, while robotic agents emulate physical actions via real-world interaction \cite{li2024embodied, cheng2024exploring}. Models such as LAMs (Large Action Models) aim to unify these capabilities by learning from action trajectories across digital and physical contexts \cite{singhal2024large}. Crucially, just as humans utilize tools to extend cognitive and physical abilities, AI agents incorporate external APIs, robotic systems, or software interfaces to achieve complex tasks \cite{angulo2023towards}. These tool-mediated actions expand the agent’s action space, mirroring the human capacity for tool use and enabling more generalized problem-solving capabilities.

\subsubsection{World Models: Cognitive Foundations Bridging Human and AGI}

World models are internal representations that allow agents to simulate, predict, and plan without depending solely on trial-and-error~\cite{hu2023language}. In humans, these mental models underpin spatial navigation, planning, and counterfactual reasoning~\cite{johnson2010mental}, offering predictive, adaptive, and scalable cognition~\cite{suomala2022human}. For instance, crossing a busy street involves anticipating vehicle motion, timing decisions, and dynamically adjusting behavior hallmarks of world model reasoning.
Figure~\ref{fig:Illustrationvisualization} illustrates the cognitive pipeline shared by human and artificial intelligence using the example of a soccer player (AI-generated \textbf{Lionel Messi}) predicting and striking a ball. The scenario demonstrates how internal world models enable trajectory prediction before motor action. Prediction integrates visual cues and prior experience, refined by perception and memory. Action is selected through an AI-like decision-making module, and feedback updates memory and internal models. The figure is structured across four conceptual layers: (1) foundational world model types (implicit, explicit, simulator-based, instruction-driven); (2) dynamic reasoning via prediction, hierarchy, and feedback; (3) core agentic faculties perception, memory, and action; and (4) aspirational AGI capabilities including ethical reasoning and contextual adaptability.

\begin{table*}[ht]
\centering
\caption{Mapping of human brain regions to neural network models and their functional parallels in AGI research.}
\footnotesize
\resizebox{\textwidth}{!}{
\begin{tabular}{p{2.5cm}p{3cm}p{3cm} p{3cm} p{4.4cm}}
\toprule
\rowcolor{lightgray}
\textbf{Brain Region / Function} & \textbf{Cognitive Role} & \textbf{Neural Network Model} & \textbf{Application} & \textbf{Comparison Highlight} \\
\midrule

Occipital Lobe & Visual processing & Convolutional Neural Networks (CNNs) & Image recognition, object detection & Biological vision uses sparse, hierarchical filtering; CNNs apply layered filters for edges and textures \\

Hippocampus / Temporal Lobe & Memory encoding, sequence modeling & Recurrent Neural Networks (RNNs), LSTMs & Sequential modeling, time-series prediction & Humans recall context adaptively; RNNs capture limited temporal state \\

Motor Cortex & Voluntary motion control & Robotic Control Networks & Robotics, motor skill learning & Human motion uses proprioception and feedback; robotic policies rely on optimization \\

Prefrontal Cortex & Planning and decision making & Reinforcement Learning (RL) & Game playing, navigation, strategy tasks & Humans plan under uncertainty and values; RL focuses on reward maximization \\

Synaptic Plasticity & Learning through temporal dynamics & Spiking Neural Networks (SNNs) & Neuromorphic modeling, real-time inference & Hebbian/STDP rules guide human learning; SNNs simulate spikes with scalability trade-offs \\

Auditory Cortex & Language and speech understanding & Transformer Networks & Language modeling, translation, text generation & Humans integrate emotion and context; Transformers use token attention over sequences \\

\bottomrule
\end{tabular}
}
\label{tab:brain-nn-comparison}
\end{table*}

\subsubsection{Neural Networks Inspired by Brain Functions}

Biological neural systems have inspired a range of architectures that replicate human cognitive functions. Convolutional Neural Networks (CNNs) and attention-based models emulate the visual cortex, excelling in learning local and global patterns~\cite{zafar2024single}. Recurrent Neural Networks (RNNs), reflecting hippocampal temporal processing, are well-suited for sequential data and memory tasks. Spiking Neural Networks (SNNs) mimic neural dynamics like synaptic plasticity and spike timing, offering advantages for temporal modeling and sensor data. Reinforcement Learning (RL), modeled on prefrontal decision-making, enables agents to learn from interaction and feedback in complex environments. Table~\ref{tab:brain-nn-comparison} summarizes how human brain regions map to neural network architectures, outlining their cognitive functions, AI analogues, and applications.
\subsection{Cognitive Processes}
Cognitive neuroscience leverages brain mapping techniques such as Electroencephalography (EEG), Electrocorticography (ECoG), Magnetoencephalography (MEG), Functional Magnetic Resonance Imaging (fMRI), and Positron Emission Tomography (PET) to investigate the neural basis of cognition~\cite{mcintosh1999mapping, lee2007neuroimaging}. These techniques capture neural activity in response to stimuli, revealing inter-regional communication patterns essential for cognitive functions such as memory~\cite{Legon_2016}, learning~\cite{Kao_2020}, language~\cite{Ardila_2016}, cognitive control~\cite{Spielberg_2015}, reward processing~\cite{Salimpoor_2013}, and moral reasoning~\cite{Hopp_2023, Fornito_2015}.
Furthermore, understanding how neurons communicate sheds light on the foundations of intelligence. Cognitive processes emerge from dynamic interactions across distributed brain regions~\cite{bassett2011understanding}. By linking neural activity to behavior, cognitive neuroscience bridges low-level circuitry and higher-order cognition~\cite{amunts2022linking}, offering insights for developing AI systems that emulate the integrative, adaptive capabilities of the human brain~\cite{grossberg2020toward, hu2021neuroscience}.%\begin{table}[htbp]
\subsubsection{Network Perspective of the Brain}
The brain functions as a complex biological network orchestrating perception, emotion, and cognition~\cite{Kriegeskorte_Douglas_2018, Park_Friston_2013}. Advances in neuroimaging and network science have enabled mapping of the brain’s structural and functional connectivity known as the connectome revealing its hierarchical and modular organization~\cite{Fornito_2016, Sporns_2013}. Brain networks are typically classified into three types: anatomical (physical infrastructure ), functional (statistical dependencies ), and effective (causal influence)\cite{Papo_2022}. While anatomical networks change slowly, functional and effective networks are dynamic and context-dependent\cite{vandenHeuvel_Sporns_2011}, offering critical insights into cognition and adaptive behavior.
\subsubsection{Brain Networks in Cognitive Neuroscience}
Research shows that cognitive functions attention, memory, decision-making emerge from dynamic interactions across brain networks~\cite{Sporns_Betzel_2016, Medaglia_2015, Pessoa_2014}. Higher cognitive performance correlates with efficient network properties, including high global integration and short path lengths~\cite{Jung_Haier_2007, Hilger_2017}, while reduced integration is linked to cognitive decline~\cite{Farahani_2019}. This supports the view that cognitive capacity depends on the structural and functional organization of brain networks.

\subsubsection{Brain Networks Integration and AGI}
Adaptive cognition arises from flexible integration across brain modules. The frontoparietal network (FPN), for instance, dynamically routes information to support diverse cognitive demands~\cite{Cole_2013, Deco_2021}. Analogously, AGI may benefit from architectures that mirror this modular integration. A central hub coordinating specialized AI modules akin to the FPN enables dynamic reconfiguration and task-specific generalization, essential for human-level intelligence.
% \begin{tcolorbox}[colback=blue!5, colframe=blue!75!black, title=Key Insight – From Brain Networks to AGI Architecture]
\begin{tcolorbox}[colback=mybluebg, colframe=myclueborder, colbacktitle=mybluetitle, coltitle=black, fonttitle=\bfseries, top=2pt, bottom=2pt, title=Key Insight – From Brain Networks to AGI Architecture]
Cognitive neuroscience reveals that intelligence arises from dynamic, flexible integration between brain networks. Translating these principles into AGI design via hybrid architectures, modular agents, and adaptive control hubs could enable machines to emulate human-like flexibility, reasoning, and learning.
\end{tcolorbox}

\subsubsection{Bridging Biological and Artificial Systems}
AGI design must integrate symbolic reasoning with neural adaptability. While symbolic AI offers logical precision, it lacks flexibility. Conversely, neural networks excel at perception and pattern learning but lack interpretability~\cite{raza2025responsible}. Hybrid neuro-symbolic systems bridge this gap~\cite{goertzel2014artificial}. Innovations like Physics-Informed Neural Networks (PINNs)\cite{raissi2019physics} and Kolmogorov–Arnold Networks (KANs)\cite{liu2024kan} exemplify architectures that embed domain knowledge into learning, improving generalization and robustness. These methods advance AGI by fusing logic, memory, and adaptivity.

\section{Models of Machine Intelligence} \label{MImodels}

Computational Intelligence (CI) encompasses a spectrum of machine learning frameworks aimed at endowing machines with cognitive capabilities comparable to humans~\cite{das2010computational}. Bridging inspiration from biological cognition and computational abstraction, CI integrates connectionist, symbolic, and hybrid models to support reasoning, learning, perception, and decision-making cornerstones of AGI development.
\subsection{Learning Paradigms}

Modern AI systems draw on a diverse suite of learning paradigms tailored to support generalization across tasks and domains. At the foundation lie supervised and unsupervised learning: the former relies on labeled examples to learn explicit mappings, while the latter uncovers latent structures from unannotated data~\cite{alloghani2020systematic}. Semi‑supervised approaches combine scarce labeled data with abundant unlabeled samples to enhance representational quality. Self‑supervised methods including pretext tasks~\cite{babu2021xls} and contrastive learning refine feature embeddings by optimizing similarity–dissimilarity relations between input pairs.

To further boost adaptability, transfer learning enables knowledge acquired in one domain to expedite learning in related tasks~\cite{weiss2016survey}, while meta‑learning and continual learning allow rapid generalization and lifelong skill acquisition without catastrophic forgetting~\cite{fallah2020convergence}. Reinforcement learning (RL) trains agents through trial‑and‑error interaction with dynamic environments~\cite{bakker2004hierarchical}. Recent RL variants such as Learning to Think (L2T) introduce process-level, information-theoretic rewards that improve sample efficiency and general reasoning without task-specific annotations~\cite{wang2025learning}.

In AGI contexts, few-shot and zero-shot learning have emerged as essential capabilities for generalization from minimal supervision~\cite{Parnami2022}. Multi-task and multimodal learning further enable cross-domain and cross-modal abstraction \cite{raza2025humanibench}, while curriculum learning emulates human cognitive development through progressive task complexity~\cite{radford2021learning}. Shortcut learning remains a cautionary lens, highlighting how models may exploit spurious cues instead of learning robust, generalizable patterns~\cite{geirhos2020shortcut}.

\subsubsection{Representation Learning and Knowledge Transfer}

\begin{figure}[h!]
\centering
\includegraphics[width=\columnwidth]{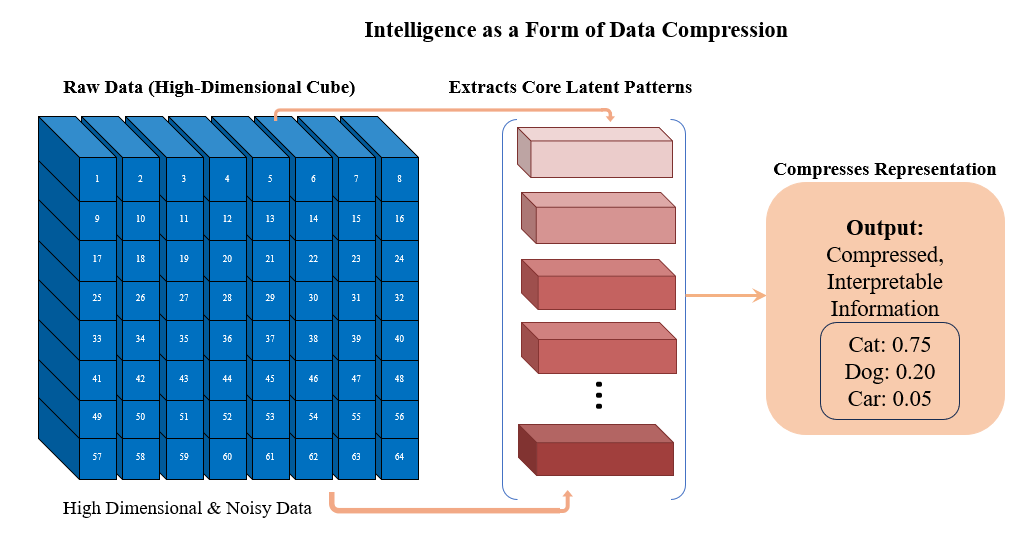}
\caption{Illustration of intelligence as compression: noisy input (left) is distilled into latent abstractions (middle) and simplified outputs (right), enhancing generalization and reasoning.}
\label{fig:intelligence}
\end{figure}

At the heart of these paradigms lies representation learning the process by which models compress raw data into compact, task-relevant abstractions. Neural networks inherently perform this compression, enabling robust transfer across tasks. As shown in Figure~\ref{fig:intelligence}, this mirrors the human brain’s ability to encode generalized, symbolic concepts rather than raw sensory inputs~\cite{higgins2022symmetry}.  Recent work~\cite{shani2025tokens} on compression–meaning tradeoffs suggests that LLMs often favor lossy statistical compression over semantic abstraction, casting doubt on their capacity for true understanding or generalization. Such compact compositional representations support adaptation, planning, and abstraction core ingredients for building versatile AGI systems.

%\subsubsection{Representation Learning and Knowledge Transfer}

%Neural networks inherently learn by compressing high-dimensional inputs into robust feature representations suited for downstream tasks, as illustrated in Fig.\ref{fig:intelligence}. Similarly, the human brain encodes generalized, reusable concepts rather than storing raw sensory inputs\cite{higgins2022symmetry}.
%Recent work on compression–meaning tradeoffs~\cite{shani2025tokens} suggests that large language models (LLMs) often prioritize lossy statistical compression over semantic abstraction, raising concerns about their capacity for true understanding and generalization. Nevertheless, these compressed internal representations whether in biological or artificial systems form the basis for adaptation, planning, and generalization across diverse and novel contexts.
\subsubsection{Knowledge Distillation} Knowledge distillation is a model optimization technique that enables the transfer of capabilities from large teacher models to smaller student models, preserving performance while improving efficiency crucial for scalable AGI systems~\cite{hinton2015distilling}. Distillation can be feature-based (aligning internal representations), response-based (matching output distributions), or relation-based (preserving structural dependencies). Variants like self-distillation, online distillation, and quantized distillation support continual learning and deployment in resource-constrained AGI environments.

% \begin{tcolorbox}[colback=blue!5!white, colframe=blue!75!black, title=Intelligence as a form of learning compressed representation and]
\begin{tcolorbox}[colback=mybluebg, colframe=myclueborder, colbacktitle=mybluetitle, coltitle=black, fonttitle=\bfseries, top=2pt, bottom=2pt, title=Intelligence as a form of learning compressed representation]
Intelligence can also be viewed as the capacity to compress high-dimensional data into abstract, low-dimensional representations~\cite{huang2024compression}. This process involves extracting structure, eliminating redundancy, and preserving key patterns for reasoning and generalization. 
\end{tcolorbox}

\subsection{Biologically and Physically Inspired Architectures}

Below, we discuss biologically and physically inspired neural architectures.

\textbf{Spiking Neural Networks (SNNs)} emulate neural spike dynamics and are ideal for temporal and event-based processing~\cite{wang2020supervised}. Their biological plausibility supports neuromorphic computing and sensorimotor control.

\textbf{Physics-Informed Neural Networks (PINNs)} incorporate physical laws (e.g., Partial Diffrential Equations (PDEs)) into neural architectures~\cite{raissi2019physics}, ensuring consistency with real-world constraints in domains such as fluid dynamics and biomechanics.

\textbf{Kolmogorov-Arnold Networks} Kolmogorov-Arnold Networks (KANs)~\cite{liu2024kan} use learnable spline-based activation functions rather than fixed ones to model complex functions, shifting the learning emphasis from weights to activations. This enhances interpretability and flexibility but requires careful regularization for stable training. Table~\ref{tab:agi-arch-theory} and \ref{tab:agi-opt-priors} summarizes the comparative strengths of SNNs, PINNs, and conventional neural networks across key AGI-relevant dimensions, including time modeling, biological plausibility, efficiency, and application scope.

\subsubsection{Symbolic, Connectionist, and Hybrid Systems}
\textbf{Symbolic AI}~\cite{sheth2024neurosymbolic} excels in interpretability and rule-based reasoning but lacks robustness in perception. \textbf{Connectionist models}~\cite{smolensky1987connectionist} (e.g., neural networks) offer scalable pattern recognition with less interpretability. Their fusion in \textbf{hybrid systems}~\cite{chahkoutahi2017seasonal} integrates structured reasoning with perceptual learning making them strong candidates for AGI architectures.

\vspace{0.2em}
% \begin{tcolorbox}[colback=blue!5!white, colframe=blue!75!black, title=Key Insight: Toward Cognitive Foundations for AGI]
\begin{tcolorbox}[colback=mybluebg, colframe=myclueborder, colbacktitle=mybluetitle, coltitle=black, fonttitle=\bfseries, top=2pt, bottom=2pt, title=Key Insight: Toward Cognitive Foundations for AGI]
  The convergence of biologically plausible dynamics (SNNs), physically constrained reasoning (PINNs), symbolic-connectionist hybrids, and advanced learning paradigms marks a decisive step toward AGI. These models enable grounded abstraction, multi-task learning, and flexible adaptation beyond pattern recognition.
\end{tcolorbox}

%\begin{table*}[htbp]
%\centering
  %\caption{Comparison of Spiking Neural Networks (SNNs), Physics-Informed Neural Networks (PINNs), and Conventional Neural Networks (NNs) across AGI-relevant dimensions.}
  %\label{tab:nn-comparison}
 % \small
  %\begin{tabular}{p{3cm}p{3.5cm}p{3.5cm}p{3.5cm}}
   % \toprule
    %\textbf{Aspect} & \textbf{SNNs} & \textbf{PINNs} & \textbf{Conventional NNs} \\
    %\midrule
    %Neuron Model & Spiking dynamics mimic biological neurons & Physics-constrained artificial neurons & Feedforward artificial neurons \\
    %Time Representation & Explicit via spike timing & Implicit or absent & Absent \\
   % Information Processing & Temporal spike patterns & Physics-informed inference & Data-driven weight learning \\
  %  Biological Plausibility & High & Moderate & Low \\
 %   Computational Efficiency & Low–Moderate (temporal overhead) & Task-dependent & High (GPU-optimized) \\
  %  Applications & Robotics, neuromorphic systems & Scientific computing, PDE solving & NLP, vision, speech \\
 %   AGI Relevance & Sensorimotor grounding, efficiency & Physical realism & Scale and pattern abstraction \\
  %  \bottomrule
 % \end{tabular}
%\end{table*}

\subsection{Intelligence as Meta-Heuristics}
General intelligence can be viewed as a dynamic collection of meta-heuristics and adaptive strategies that continuously evaluate, revise, and optimize problem-solving pathways~\cite{benderskaya2013multidisciplinary}. Unlike fixed heuristics~\cite{silver2004overview}, meta-heuristic agents improve iteratively by learning from failure and adapting strategies across domains. Recent AGI frameworks such as  AutoGPT \cite{yang2023auto}, and Voyager \cite{wang2023voyager} demonstrate such behavior through internal feedback loops, self-prompting, and chain-of-thought reasoning. These systems optimize both task-specific performance and the broader process of learning itself, supporting transfer, adaptability, and generalization~\cite{zou2024review}. Intelligence, in this view, is not a static capacity but a recursive, self-improving search over heuristics.

\subsection{Explainable AI (XAI)}
As AI advances toward AGI, explainability must evolve from post hoc interpretation to intrinsic transparency. Traditional techniques such as saliency maps and Grad-CAM provide limited insights into model reasoning~\cite{rane2024explainable, neri2023explainable}. AGI systems, however, require explainability that mirrors human cognition enabling agents to articulate not just outcomes, but the rationale behind decisions~\cite{rai2020explainable}.

This calls for architectural integration of interpretability through neuro-symbolic reasoning~\cite{lu2024surveying}, causal modeling~\cite{bengio2019meta}, and biologically inspired mechanisms such as memory traces and attention routing. 
 Furthermore, multi-level explanations tailored to diverse user contexts are essential~\cite{raza2025responsible,raza2025just}. Embedding meta-cognition and self-aware justification as core design principles will transform XAI from an afterthought to a foundational component of general intelligence.
\section{Generalization in Deep Learning}
\label{{GenDL}}

Generalization in deep learning refers to a model’s ability to extend learned patterns from training data to unseen scenarios, making it essential for AGI development~\cite{kawaguchi2017generalization}. Unlike narrow AI, which often overfits task-specific distributions, AGI systems must demonstrate robust transferability across domains and contexts~\cite{na2024emergence}.

\begin{table*}[htbp]
\centering
\small
\caption{
Architectures and Generalization Theory in AGI: (A) neuro-inspired and physics-informed designs (e.g., SNNs, PINNs); (B) theoretical constructs (e.g., IB, MDL, NTK).
}
\label{tab:agi-arch-theory}
\begin{adjustbox}{max width=\textwidth}
\footnotesize
\begin{tabularx}{\textwidth}{l X X X}
\toprule
\rowcolor{lightgray}
\multicolumn{4}{c}{\textbf{Panel A: Neuro-Inspired and Physics-Grounded Architectures}} \\
\toprule
\textbf{Architectures} & \textbf{SNNs} & \textbf{PINNs} & \textbf{Conventional NNs} \\
\midrule
Property & Simulate spike-timing and event-driven signaling & Encode physical constraints within neural units & Abstract artificial neurons using trainable weights \\
Time Dynamics & Temporal encoding via spikes & Task-driven implicit time representation & Often absent unless RNNs are used \\
Computation Paradigm & Event-based, energy-efficient processing & PDE-constrained data fitting & Data-driven general-purpose mapping \\
Biological Alignment & High (plasticity, sparsity) & Moderate (physics realism) & Low (flexible but abstract) \\
Efficiency & Moderate; optimized  & Dependent on solver complexity & High throughput/GPU parallelism \\
Use Cases & Edge robotics, dynamic sensing & Scientific simulation, climate modeling & Vision, NLP, reinforcement learning \\
AGI Potential & Real-time perception & Symbol grounding via physics & Scalable pattern abstraction \\
\bottomrule
\end{tabularx}
\end{adjustbox}

\vspace{0.5em}

\begin{adjustbox}{max width=\textwidth}
\small

\begin{tabularx}{\textwidth}{l X X X}
\rowcolor{lightgray}
\multicolumn{4}{c}{\textbf{Panel B: Theoretical Constructs for Generalization}} \\
\toprule
\textbf{Theory} & \textbf{Inductive Principle} & \textbf{Foundation} & \textbf{Implication for AGI} \\
\midrule
Information Bottleneck (IB) & Focus on relevant latent features while discarding noise & Information theory, mutual information & Compact, task-relevant representation learning \\
Minimum Description Length (MDL) & Simplicity favors generalization & Algorithmic info theory & Selects compressed, interpretable models \\
Implicit Regularization (SGD) & Flat minima during optimization & Loss landscape geometry &  Encourages generalization \\
NTK / Double Descent & Overparameterized regimes benefit late generalization & Infinite-width kernel theory & Characterizes regimes of robust learning \\
PAC-Bayes Bounds & Generalization from distributional priors & Probabilistic learning theory & Formal generalization guarantees \\
Causal Representation Learning & Extracts stable causal features invariant to interventions & Causal graphs, SEMs & Promotes robustness across tasks/distributions \\
Variational Dropout & Regularizes through learned noise injection & Variational inference & Enforces sparsity and noise resilience \\
Simplicity Bias & Learns simpler hypotheses first & Empirical dynamics of training & Lower complexity early in training \\
\bottomrule
\end{tabularx}
\end{adjustbox}
\end{table*}

\begin{table*}[htbp]
\centering
\caption{
Optimization and Priors in AGI: (C) learning algorithm biases (e.g., SGD, RL, PEFT); (D) emerging priors in foundation models (e.g., RAG, MAE, RLHF).
}
\label{tab:agi-opt-priors}
\begin{adjustbox}{max width=\textwidth}
\footnotesize
\begin{tabularx}{\textwidth}{l X X X}
\rowcolor{lightgray}
\multicolumn{4}{c}{\textbf{Panel C: Learning Algorithms and Loss Function Biases}} \\
\toprule
\textbf{Mechanism} & \textbf{Inductive Bias} & \textbf{Examples} & \textbf{Relevance to AGI} \\
\midrule
SGD / Early Stopping & Implicit preference for flatter minima & Classic training setups & Generalizable, stable convergence \\
Adaptive Optimizers (Adam, RMSProp) & Faster convergence but risk of sharp solutions & LLM fine-tuning, low-data setups & Tradeoff between speed and generalization \\
Cross-Entropy Loss & Promotes confident predictions & Classification tasks & Simple yet insensitive to uncertainty \\
Contrastive / Triplet Loss & Latent clustering, relational structure & SimCLR, MoCo, triplet nets & Robust representation learning \\
KL Divergence (in VAEs, PAC-Bayes) & Regularizes latent space or distributions & VIB, Bayesian networks & Encourages minimal, disentangled codes \\
RL Objectives & Long-term credit assignment, goal focus & PPO, Q-learning, DPO & Supports planning and sequential reasoning \\
Meta-Learning / PEFT & Task-agnostic initialization or fast adaptation & MAML, LoRA, Reptile & Enables efficient few-shot or continual learning \\
\bottomrule
\end{tabularx}
\end{adjustbox}

\vspace{0.5em}

\begin{adjustbox}{max width=\textwidth}
\footnotesize
\begin{tabularx}{\textwidth}{l X X X}
\rowcolor{lightgray}
\multicolumn{4}{c}{\textbf{Panel D: Emerging Inductive Priors in Foundation Models}} \\
\toprule
\textbf{Mechanism} & \textbf{Inductive Bias} & \textbf{Examples} & \textbf{AGI Relevance} \\
\midrule
Multimodal Attention & Enables alignment across modalities & CLIP, Flamingo, Perceiver IO & Supports grounded reasoning and perceptual understanding \\
Cross-Modal Contrastive Learning & Aligns visual and language embeddings via shared structure & ALIGN, LiT, GIT & Encourages shared representations and compositionality \\
External Memory Augmentation & Facilitates long-term and episodic recall & RNN+Memory, ReAct, RETRO & Enables scalable context and symbolic chaining \\
Retrieval-Augmented Generation (RAG) & External database during inference & RAG, Atlas, KAT & Enhances factuality/adaptability \\
Masked Modeling / Autoregression & Learns predictive structure from partial context & BERT, GPT, BEiT, MAE & General-purpose self-supervised pretraining \\
Prompt Tuning and Instruction Biases & Learns structure through task prompts or instructions & T5, InstructGPT, PEFT, Prefix Tuning & Provides zero-shot adaptation and alignment with user intent \\
RL with Human Feedback (RLHF) & Aligns model outputs with human values/preferences & InstructGPT, DPO, Constitutional AI & Critical for safety and value alignment \\
\bottomrule
\end{tabularx}
\end{adjustbox}
\end{table*}

\subsection{Foundations of Generalization in AGI}
Robust generalization is a cornerstone of AGI, enabling systems to adapt beyond their training distribution. Let $P$ represent the training data distribution and $Q$ the real-world distribution. The empirical risk $R_{\text{emp}}$ measures training error, while $R_{\text{general}}$ reflects expected real-world error. The generalization gap $R_{\text{emp}} - R_{\text{general}}$ captures how well a model extrapolates to new settings. A strong and robust AGI system should have lower generalization gaps.
Theoretical frameworks have highlighted several different perspectives of generalization as follows.

\textbf{Information Bottleneck (IB)} theory proposes that models generalize by compressing inputs into compact latent representations that preserve only task-relevant information while discarding irrelevant or spurious signals~\cite{tishby2000information}. This compression principle provides a trade-off between retaining predictive power and limiting unnecessary input information, thereby constraining model complexity. Shwartz-Ziv and Tishby~\cite{shwartz2017opening} were among the first to empirically and theoretically propose that deep neural networks progressively compress representations as they learn, connecting this to improved generalization. Their follow-up work with Painsky~\cite{shwartz2018representation} offered further theoretical support and a sample-complexity-oriented bound linking information compression to generalization. Building on these ideas, Kawaguchi et al.~\cite{kawaguchi2017generalization} later developed rigorous statistical learning bounds formalizing this principle in modern deep architectures. More recently, Shwartz-Ziv and LeCun~\cite{shwartz2024toCompress} extended these information bottleneck arguments to the self-supervised learning paradigm, suggesting that compression not only benefits supervised generalization but also plays a key role in representation learning without labels. This sequence of work suggests that the information bottleneck is not only cognitively and biologically plausible but also grounded in solid mathematical and empirical evidence.

\textbf{Minimum Description Length (MDL)} is based on the idea that the simplest explanation or model that best compresses the data will generalize better~\cite{schmidhuber2007simple}. MDL suggests that simpler models, which can compress data better, are less likely to overfit and thus generalize more effectively.

\textbf{Implicit Regularization}, often associated with stochastic gradient descent (SGD), suggests that optimization methods naturally bias models toward flat minima, which stems from the geometry of loss landscapes and provides insight into how generalization arises without explicit regularization~\cite{neyshabur2014search}.

\textbf{Neural Tangent Kernel (NTK)} and \textbf{Double Descent} theories together offer a modern understanding of generalization in overparameterized neural networks. NTK shows that as network width approaches infinity, training dynamics become linear and predictable, behaving like kernel regression and often leading to well-generalizing solutions despite large model sizes~\cite{jacot2018neural}. Double Descent complements this by revealing that increasing model capacity initially leads to overfitting near the interpolation threshold, but further scaling results in a second descent in test error with improved generalization~\cite{belkin2019reconciling}.

\textbf{PAC-Bayes Bounds} combine elements of Bayesian inference with Probably Approximately Correct (PAC) learning~\cite{mcallester1999pac}. They bound the generalization error of a hypothesis based on its divergence from a prior, typically measured via the Kullback-Leibler (KL) divergence.

\textbf{Causal Representation Learning} emphasizes learning representations that capture the causal structure of data, rather than mere statistical correlations~\cite{scholkopf2021toward}. It uses tools from causal inference, such as structural equation models and do-calculus, to extract invariant features under interventions.

\textbf{Variational Dropout} is a Bayesian regularization method that interprets dropout as approximate variational inference~\cite{kingma2015variational}. It injects noise into the model’s weights using a learnable distribution, often leading to sparsity and robustness. Unlike fixed dropout rates, variational dropout adapts the noise level during training, improving generalization in uncertain or noisy environments.

\textbf{Simplicity Bias} refers to the empirical observation that deep networks, when trained with gradient descent, tend to learn simpler functions before complex ones~\cite{valle2018deep}. This bias arises from the implicit properties of parameter-function mappings and the dynamics of neural network training. As a result, models are more likely to converge to functions with lower complexity, which tend to generalize better.

% \begin{tcolorbox}[colback=blue!5!white, colframe=blue!75!black, title=Generalization: A Pillar of AGI]
\begin{tcolorbox}[colback=mybluebg, colframe=myclueborder, colbacktitle=mybluetitle, coltitle=black, fonttitle=\bfseries, top=2pt, bottom=2pt, title=Generalization: A Pillar of AGI]
Effective generalization not just memorization distinguishes AGI from narrow AI. Theories like the Information Bottleneck, minimum description length, and optimization landscapes converge on one idea: compress inputs to extract robust, transferable representations.
\end{tcolorbox}

\subsection{Architectural and Algorithmic Inductive Biases}
Inductive biases embedded in model architectures and learning algorithms are central to the design of AGI systems, guiding how they learn, generalize, and reason. For example, linear models offer interpretability but are limited in capturing nonlinear patterns~\cite{huang2024compression}. MLPs support hierarchical representations but lack spatial or temporal priors~\cite{popescu2009multilayer}. CNNs introduce local spatial bias and translation invariance ideal for vision while RNNs model sequences but struggle with long-range dependencies~\cite{sherstinsky2020fundamentals}. Transformers~\cite{vaswani2017attention}, with global attention, excel at long-range modeling and underpin modern LLMs like GPT~\cite{lu2024unified}, though they lack grounded abstraction. State-space models (e.g., Mamba) offer implicit recurrence and dynamic memory~\cite{gu2023mamba}, improving temporal scalability. GNNs encode relational priors for graph-structured tasks~\cite{battaglia2018relational}, and GANs~\cite{goodfellow2014generative} support powerful generative modeling, albeit with stability trade-offs.

\subsubsection{Biases in Learning Algorithms} Learning algorithm biases also play a vital role. Optimization methods like SGD favor flat minima with better generalization~\cite{xie2020diffusion}, while adaptive optimizers like Adam can converge faster but bias toward sharper solutions~\cite{de2018convergence}. Loss functions impose task-specific priors: cross-entropy for classification, contrastive losses for relational tasks, and adversarial or reinforcement losses for realism and long-term planning~\cite{wang2022comprehensive}. Meta-learning and structured losses promote compositionality and generalization across tasks essential traits for AGI. A unified AGI architecture may need to integrate these diverse inductive structures to achieve abstraction, compositionality, and adaptive reasoning across modalities and tasks.

\subsubsection{Solving Inductive Bias Technique}

AGI systems must generalize not only across tasks but also across distributions, time, and embodiment. Techniques to enhance this capability include \textbf{uncertainty estimation}, which accounts for epistemic and aleatoric uncertainty to improve reliability~\cite{sedlmeier2021quantifying} (further discussed in Section X), and
 \textbf{adaptive regularization} mitigates catastrophic forgetting in continual learning~\cite{anil2019memory}.
   
\subsection{Generalization During Deployment}

% \paragraph{Test-Time Adaptation (TTA):} Allows models to update representations dynamically at inference~\cite{liang2025comprehensive}.
\textbf{Test-Time Adaptation (TTA)} refers to techniques that enable machine learning models to dynamically adjust their predictions at inference time, aiming to improve robustness to distributional shifts or domain changes encountered during deployment~\cite{liang2025comprehensive}. There are two primary paradigms within TTA: optimization-based TTA and training-free TTA.
\paragraph{Optimization-based TTA} involves updating certain model parameters, typically through gradient descent, at test time, using unsupervised or self-supervised objectives derived from the test data itself, such as test-time training (TTT)~\cite{sun2020test} and test-time prompt tuning (TPT)~\cite{shu2022test}. 
\paragraph{Training-free TTA} improves model adaptation at test time without performing any explicit parameter updates or gradient-based optimization. Instead, these methods rely on recalibrating or modifying the model’s inference process, such as training-free dynamic adapter (TDA)~\cite{karmanov2024efficient} and dual memory network (DMN)~\cite{zhang2024dual}

% \paragraph{Test-Time Training (TTT):} Fine-tunes parameters in response to real-world drift using incoming data streams.

\textbf{Retrieval-Augmented Generation (RAG)} augments model predictions by incorporating information retrieved from large external databases, document corpora, or knowledge bases during inference~\cite{lewis2020retrieval, shi2023replug}. Instead of relying solely on the parametric memory of the model, RAG retrieves relevant documents or facts in response to a query or input and conditions the model’s output on both the original input and the retrieved evidence. RAG can improve factual accuracy and reduce hallucination without requiring additional model retraining, but challenges include efficient retrieval, handling noisy evidence and latency during inference.

% \begin{tcolorbox}[colback=blue!5!white, colframe=blue!75!black, title=Deployment-Time Generalization]

\begin{tcolorbox}[colback=mybluebg, colframe=myclueborder, colbacktitle=mybluetitle, coltitle=black, fonttitle=\bfseries, top=2pt, bottom=2pt, title=Deployment-Time Generalization]
For AGI to succeed in dynamic environments, continual adaptation is essential. Techniques like TTA and RAG offer real-time resilience through knowledge retrieval, error correction, and ongoing learning.
\end{tcolorbox}

\subsection{Toward Real-World Adaptation}

\paragraph{Embodied Intelligence} To achieve real-world adaptation, AGI systems must bridge the gap between abstract reasoning and physical interaction. This requires the integration of perception, planning, and control to enable flexible behavior in dynamic environments. Techniques such as imitation learning and zero-shot planning are instrumental for equipping robots and embodied agents with the ability to generalize learned knowledge to novel tasks and contexts, thereby enhancing adaptability and autonomy in robotics applications~\cite{paolo2024call}.
% AGI must translate abstract reasoning into physical interaction. Imitation learning and zero-shot planning enable adaptability in robotics~\cite{paolo2024call}.

\paragraph{Causal Reasoning} Robust adaptation necessitates distinguishing causation from mere correlation, a challenge addressed by the causal inference frameworks pioneered by Pearl and Bengio~\cite{bengio2019meta}. Causal reasoning allows AGI to identify and model underlying mechanisms, supporting effective generalization across distribution shifts and facilitating reliable interventions in complex, uncertain environments. 
% Pearl and Bengio’s causal models help distinguish correlation from causation, crucial for domain shifts and interventions~\cite{bengio2019meta}.

\paragraph{Robustness and Alignment} AGI must be resilient to rare, high-impact "black swan" events that are difficult to anticipate but potentially catastrophic. Ensuring robustness involves the capacity for safe exploration, rapid adaptation to unforeseen scenarios, and continual monitoring for emergent risks. At the same time, alignment mechanisms are critical to guarantee that AGI systems consistently act in accordance with human values and intentions, even in the face of novel and ambiguous circumstances~\cite{ngo2022alignment}.

\section{Reinforcement Learning and Alignment for AGI}

\textit{“The measure of intelligence is the ability to change”} \textbf{(Albert Einstein)}. This insight underscores a limitation of static neural networks: true intelligence demands adaptability. Reinforcement learning (RL), which enables agents to learn by interacting with their environment and adapting through feedback, captures this essence~\cite{sutton1998reinforcement, silver2017mastering}. Unlike supervised learning, which relies on fixed datasets, RL thrives in non-stationary, uncertain environments, making it a natural candidate for AGI~\cite{haarnoja2018soft}.

% \begin{tcolorbox}[colback=blue!5!white, colframe=blue!75!black, title=The Core of AGI: Learning by Doing in Real Time]
\begin{tcolorbox}[colback=mybluebg, colframe=myclueborder, colbacktitle=mybluetitle, coltitle=black, fonttitle=\bfseries, top=2pt, bottom=2pt, title=The Core of AGI: Learning by Doing in Real -Time]
RL’s foundation lies in its trial-and-error paradigm, promoting continual, adaptive learning through experience.
\end{tcolorbox}

\subsection{Reinforcement Learning: Cognitive Foundations}

While RL offers a promising path toward adaptive intelligence, its direct application to AGI is hindered by several limitations, including sample inefficiency, limited scalability in high-dimensional spaces, and vulnerability to reward misspecification~\cite{haarnoja2018soft, everitt2018agi}. To address these concerns, algorithmic strategies have been developed.

\textbf{Model-based RL} incorporates predictive dynamics to reduce sample complexity~\cite{silver2017mastering}, while \textbf{hierarchical RL} decomposes tasks into reusable subtasks for more efficient exploration and planning~\cite{bakker2004hierarchical}. Complementing these advances, cognitive reasoning methods inspired by LLMs significantly expand RL's expressive capacity.

Recent methods such as \textbf{Chain-of-Thought (CoT)}~\cite{wei2022chain}, \textbf{Tree-of-Thought (ToT)}~\cite{yao2023tree}, and \textbf{Reasoning-Acting (ReAct)}~\cite{yao2023react} embed structured, deliberative reasoning within RL pipelines. CoT enables transparent multi-step inference; ToT explores multiple solution paths to improve policy selection; and ReAct integrates reasoning with environment interaction, reducing errors and enhancing adaptability. These methods mitigate short-term bias and inefficient exploration, aligning RL agents more closely with the demands of general intelligence~\cite{shakya2023reinforcement}.

\textbf{Integrative frameworks} exemplify this convergence of RL and LLM reasoning:
\begin{itemize}
    \item \textbf{MetaGPT}~\cite{hong2023metagpt}: Coordinates multiple LLM agents in specialized roles, facilitating structured task decomposition and collaborative problem-solving.
    \item \textbf{SwarmGPT}~\cite{jiao2023swarm}: Combines LLM planning with multi-agent RL for real-time coordination in systems such as robotic swarms.
    \item \textbf{AutoGPT}~\cite{yang2023auto}: Demonstrates autonomous goal decomposition, iterative self-correction, and continuous self-improvement via internal RL loops.
\end{itemize}
Supporting these frameworks are optimization strategies such as:
\begin{itemize}
    \item \textbf{Proximal Policy Optimization (PPO)}~\cite{schulman2017proximal}: Balances policy performance with stability.
    \item \textbf{Direct Preference Optimization (DPO)}~\cite{rafailov2023direct}: Trains agents directly from preference data, simplifying alignment.
    \item \textbf{Group Relative Policy Optimization (GRPO)}~\cite{shinn2023reflexion}: Optimizes reasoning quality by comparing multiple generated trajectories.
\end{itemize}

\subsection{Human Feedback and Alignment}
\textbf{Reinforcement Learning with Human Feedback (RLHF)}~\cite{ouyang2022training} addresses AGI alignment by incorporating human judgments into the reward loop, improving safety and reducing harmful outputs~\cite{christiano2017deep, bai2022training}. RLHF underpins systems like InstructGPT and ChatGPT, though challenges remain in scaling feedback and mitigating biases.

\subsubsection{Alignment Techniques and Supervision}

\textbf{Human-in-the-loop} training, \textbf{value learning}, and \textbf{inverse reinforcement learning} enhance AGI's alignment with human values~\cite{gao2024artificial}. Online supervision allows real-time adaptation~\cite{salmon2023managing}, while offline supervision enables reflective policy refinement without continuous oversight~\cite{andreoni2024enhancing, zhao2024offline, raza2025vldbench}. Additionally, machine unlearning~\cite{liu2025rethinking} has emerged as a corrective tool for removing spurious correlations, hallucinations, or biased representations in vision-language models, contributing to safer and more interpretable systems~\cite{narnaware2025sb}.

\subsubsection{Ethical Issues of AGI}

As AGI systems approach greater autonomy and capability, ensuring fairness, transparency, trust, and privacy becomes not only a technical imperative but also a societal one~\cite{shneiderman2020bridging, khan2024agi, raza2025humanibench}. These principles form the ethical backbone of safe AGI deployment, safeguarding individuals and communities from disproportionate harms such as surveillance, exclusion, or algorithmic manipulation.
To address these challenges, governance frameworks must be grounded in human rights and international norms~\cite{mylrea2023artificial, yan2024practical}. These frameworks must go beyond technical safeguards by incorporating participatory design, redress mechanisms, and interdisciplinary oversight. Without such structures, AGI risks reinforcing existing inequities, centralizing power, and becoming unaccountable in high-stakes decisions.

\subsubsection{Future Outlook}

Future alignment strategies must integrate multidisciplinary insights from AI, ethics, psychology, and law~\cite{bikkasani2024navigating, raman2025navigating}. As shown in Figure~\ref{fig:vlm_evolution2}(a), AGI readiness hinges on cognitive, interface, systems, and alignment axes. Figure~\ref{fig:vlm_evolution2}(b) shows expert uncertainty, with 37\% expecting AGI realization in two decades or more~\cite{fengfar}. Cross-cultural modeling, robust evaluation, and international coordination will be critical.

\section{AGI Capabilities, Alignment, and Societal Integration}
AGI seeks to replicate core human cognitive abilities reasoning, learning, memory, perception, and emotion to operate autonomously across domains~\cite{fengfar}. Beyond technical capability, safe deployment requires alignment with ethical principles and social values. This section synthesizes cognitive foundations, psychological insights, and governance frameworks that shape AGI’s path toward responsible integration~\cite{mclean2023risks}.

\begin{tcolorbox}[colback=mybluebg, colframe=myclueborder, colbacktitle=mybluetitle, coltitle=black, fonttitle=\bfseries, top=2pt, bottom=2pt, title=AGI Integration at a Glance]
\label{p21}
\textbf{Cognitive Core:} Reasoning, learning, memory, and perception underpin AGI adaptability.\\
\textbf{Safety:} Robust design, value alignment, and human-in-the-loop controls remain essential.\\
\textbf{Psychological Grounding:} Cognitive science guides realistic and ethical agent behavior.\\
\textbf{Governance:} Frameworks like NIST, EU AI Act, and OECD foster transparent oversight.\\
\textbf{Equity:} “AI for everyone, by everyone” reflects the need for co-design and fair access.
\end{tcolorbox}

\subsection{Core Cognitive Functions}

\subsubsection{Reasoning} 
AGI systems must perform deductive, inductive, and abductive reasoning to solve novel problems~\cite{dwivedi2023opinion,zhao2023brain}. Deep reasoning enables hypothesis testing, planning, and counterfactual inference\cite{zhang2023critical}. Models like chain-of-thought and neuro-symbolic systems integrate symbolic logic with neural learning for more interpretable and adaptive reasoning \cite{mariani2024generative,ali2024cognitive, campos2025gaea}.
\subsubsection{Learning}
AGI integrates supervised, unsupervised, symbolic, reinforcement, and deep learning paradigms~\cite{wang2018self,lecun2015deep}. These enable generalization and continuous refinement. Reinforcement learning facilitates interaction-based learning in dynamic environments~\cite{golpayegani2024advancing}, while deep learning abstracts features across modalities~\cite{taye2023understanding}.
\subsubsection{Thinking}
Thinking refers to abstraction, strategy formation, and decision-making. Cognitive architectures and neural networks simulate high-level thought~\cite{bojic2024cern}. Neuro-symbolic systems combine formal logic with adaptable models~\cite{li2023vertical}, increasing reliability in complex reasoning tasks~\cite{mikkilineni2024digital}.

\subsubsection{Memory}
Memory supports context awareness and learning continuity. Short-term memory aids in immediate task handling; long-term memory encodes cumulative knowledge~\cite{lampinen2021towards, isaev2023memory}. Parametric and external memory systems allow rapid retrieval and flexible updates~\cite{goertzel2023generative}.
\subsubsection{Perception}
AGI perception involves multimodal sensory interpretation. CNNs and transformers process visual and auditory signals\cite{schmidhuber1997long}. Advances in multimodal models like Perceiver and Flamingo improve AGI's ability to interpret heterogeneous inputs\cite{zhang2020advances}.
\subsection{Human-Centered Foundations: Psychology and Safety in AGI Design}
The safe deployment of AGI requires more than technical ingenuity; it demands architectures informed by a realistic understanding of human cognition~\cite{everitt2018agi}. Cognitive psychology reveals mechanisms such as attention, memory consolidation, emotion regulation, and causal reasoning~\cite{posner2012cognitive, baars2005global}, which inform AGI’s design and behavior modeling. Concepts like incremental learning and theory of mind~\cite{gopnik2004theory,premack1978does} offer blueprints for developing adaptive, socially attuned agents. However, naively importing psychological concepts can introduce anthropomorphic biases or flawed heuristics~\cite{rahwan2022combining}. A human-centered AGI must be empirically grounded, cross-culturally aware, and sensitive to normative variation~\cite{salhab2024systematic}. 

Safety concerns are deeply intertwined with these human-centered foundations. AGI’s open-ended generalization capabilities heighten the risk of unintended behavior~\cite{schuett2023towards}. Key dimensions include technical robustness (resilience to adversarial inputs), specification soundness (goal alignment), and human control (corrigibility, intervenability)~\cite{he2024security}. Research in scalable oversight~\cite{cihon2024chilling}, reward modeling~\cite{gu2022review}, and uncertainty calibration~\cite{burton2024uncertainty} seeks to systematically mitigate these vulnerabilities.

Ultimately, AGI systems must not only learn, plan, and reason but also reflect, defer, and ask for help~\cite{salhab2024systematic}. Embedding interpretability, human-in-the-loop safeguards, and NSFW (Not Safe for Work) content filters~\cite{guzman2023advancing} is essential for preserving public trust. Building AGI that is intelligent, safe, and aligned begins with understanding the minds it aims to augment, not replace. Table~\ref{tab:agi-panels} outlines major evaluation benchmarks, bio-inspired system mappings, and emerging governance frameworks~\cite{raza2025responsible}.

\subsection{Societal Integration and Global Frameworks}
The transition of AGI from lab to society raises urgent questions regarding equity, human agency, and democratic oversight, as shown in the Algorithm 3. 

\textbf{Work and Autonomy:} AI is not only transforming manual labor but increasingly encroaching on cognitive, techical and emotional domains. Recent studies reveal that prolonged LLM use in educational settings leads to measurable cognitive debt, marked by reduced neural engagement, memory recall, and authorship awareness~\cite{kosmyna2025your}. \par As intelligent agents begin to mediate professional and personal routines, these shifts raise profound questions about identity, equity, and the structure of work~\cite{bikkasani2024navigating}. The World Economic Forum estimates that up to 87\% of data-driven tasks could be automated by AGI~\cite{wef}, while leading AI developers suggest that most white-collar roles are now within reach of current-state-of-the-art models. These trends underscore the urgency of designing inclusive systems and proactively reimagining labor, education, and welfare infrastructures to ensure a just transition. \\

\textbf{Public Trust} Public sentiment oscillates between promise and peril. While AGI-augmented healthcare and education spark hope, concerns about surveillance and job loss demand transparent oversight, participatory development, and community-driven evaluation~\cite{naude2020race}.

\textbf{Policy Infrastructure} Several governance frameworks are converging to guide AGI deployment. The NIST AI RMF~\cite{ai2023artificial} promotes trustworthiness through interpretability and risk mitigation. The EU AI Act enforces risk-tiered compliance in high-stakes sectors. UNESCO and OECD advocate global ethical standards rooted in inclusivity, safety, and accountability~\cite{van2023ethics}.

\textbf{AI for Everyone, by Everyone} As AGI systems become more powerful, their development must reflect diverse societal needs and values~\cite{meng2025data}. The principle of "AI for everyone, by everyone" underscores the importance of participatory design, equitable access to AI resources, and co-governance across disciplines and geographies. Open-source models, community auditing, and culturally tuned datasets are crucial to democratize AGI and avoid reinforcing power asymmetries.
\textbf{Constructive Examples} Early signs of responsible integration include AI tutors, digital mental health agents, and scientific co-reasoners~\cite{raza2024developing}. These applications demonstrate the potential of AGI to increase expertise, but also underscore the need for accountability in decision-making pipelines.

\textbf{Toward Co-Designed Futures} To ensure that AGI advances human flourishing, it must be co-developed with ethicists, legal scholars, and the public. Embedding AGI within sociotechnical ecosystems~\cite{voss2023we}, through cross-disciplinary governance, inclusive norms, and transparent validation, will be critical to building systems that are not only intelligent, but also wise~\cite{luo2023designing}.
\begin{table*}[htbp]
\centering
\small
%\label{tab3}
\renewcommand{\arraystretch}{1.2}
\caption{Panel A presents representative benchmarks for AGI evaluation. Panel B maps biologically inspired cognitive functions to vision-language and agentic AI systems. Panel C outlines global governance frameworks for safe, ethical, and equitable AGI deployment.}
\label{tab:agi-panels}
\vspace{1ex}

% Panel A
\rowcolors{1}{}{gray!10}
\begin{adjustbox}{width=\textwidth}
\begin{tabular}{@{}p{2.5cm}p{2.5cm}p{3.5cm}p{3.5cm}p{2.3cm}p{2.3cm}@{}}
\multicolumn{6}{l}{\textbf{Panel A: Representative Benchmarks for AGI Evaluation}} \\
\toprule
\textbf{Benchmark} & \textbf{Focus} & \textbf{Capabilities Tested} & \textbf{Notable Feature} & \textbf{Modality} & \textbf{Interactivity Level} \\
\midrule
BIG-Bench~\cite{srivastava2022beyond} & Language reasoning & Multitask generalization, logic, math & Human-written diverse tasks & Language & Static \\
ARC~\cite{chollet2019measure} & Abstract reasoning & Concept composition & System-2 style generalization & Visual, Symbolic & Static \\
MineDojo~\cite{fan2022minedojo} & Embodied AI & Planning, exploration & Minecraft sandbox environment & Multimodal & Interactive \\
BabyAI~\cite{chevalier2018babyai} & Language grounding & Navigation, planning & Curriculum-based instructions & Language + Embodied & Interactive \\
Agentbench~\cite{liu2023agentbench} & LLM agents & Tool use, dialogue & Multi-agent evaluation & Language + Tools & Real-time \\
AGI-Bench~\cite{tang2023agibench} & AGI evaluation & Multimodal generalization & Multi-domain tasks & Multimodal & Mixed \\
eAGI~\cite{neema2025evaluation} & Engineering cognition & Reasoning, synthesis, critique & Bloom-level tasks with structured design inputs & Text + Diagrams & Mixed \\
\bottomrule
\end{tabular}
\end{adjustbox}
\vspace{3ex}

% Panel B
\rowcolors{1}{}{gray!10}
\begin{adjustbox}{width=\textwidth}
\begin{tabular}{@{}p{2.5cm}p{2.8cm}p{2.7cm}p{2.7cm}p{2.6cm}p{2.7cm}@{}}
\multicolumn{6}{l}{\textbf{Panel B: Mapping Brain-Inspired AGI Functions to Vision-Language and Agentic AI Architectures}} \\
\toprule
\textbf{AGI Function} & \textbf{Biological Inspiration} & \textbf{VLM Representation} & \textbf{Agentic AI Mechanism} & \textbf{Development Pathway} & \textbf{Future Applications} \\
\midrule
Brain Functions & Neocortex (reasoning), Hippocampus (memory), Cerebellum (motor control) \cite{wagner2020neocortex, vann2011hippocampus} & Transformer attention modules simulating cortical modularity \cite{dai2023brain} & Autonomous agents with role-based communication and planning \cite{zhuge2023mindstorms} & Neuro-symbolic cognitive architectures unifying language and perception & Cognitive robotics, brain-inspired diagnostics, and human-AI collaboration \\

Memory Systems & Hierarchical short- and long-term memory; working memory dynamics \cite{hannula2017beyond} & In-context retrieval, memory tokens, and dynamic prompt chaining \cite{dakat2024enhancing} & Persistent memory, episodic task replay, and continual learning agents \cite{sapkota2025ai} & Meta-memory and lifelong memory consolidation frameworks & Adaptive tutoring systems, emotional-aware assistants, and digital memory augmentation \\

Action Systems & Cognitive imagination, motor planning, and physical interaction \cite{schillaci2016exploration} & Scene-grounded VLM control with vision-to-action APIs \cite{liu2025advances} & Task-specialized agents under orchestration and multi-agent tool-use \cite{acharya2025agentic} & Embodied perception-action systems in real and virtual environments & Autonomous robotics in healthcare, manufacturing, and creative co-design \\

World Modeling & Internal generative simulation, counterfactuals, predictive coding \cite{mumuni2025large,calleo2025ai} & Multimodal latent embeddings and temporal scene simulation \cite{liu2025advances} & Self-play reasoning and task generation (e.g., AZR \cite{zhao2025absolute}) with verifiable feedback \cite{chen2024s} & Causal inference and forward-planning agents for open-ended tasks & Scientific reasoning, autonomous experimentation, AGI research copilots \\
\bottomrule
\end{tabular}
\end{adjustbox}

\vspace{3ex}

% Panel C
\rowcolors{1}{}{gray!10}
\begin{adjustbox}{width=\textwidth}
\footnotesize
\begin{tabular}{@{}p{3cm}p{3cm}p{3.2cm}p{3.5cm}p{2.2cm}p{2.2cm}@{}}
\multicolumn{6}{l}{\textbf{Panel C: Societal Frameworks and Policy Instruments for AGI Deployment}} \\
\toprule
\textbf{Framework} & \textbf{Institution/Origin} & \textbf{Principles} & \textbf{Key Areas Addressed} & \textbf{Scope} & \textbf{Enforcement Strategy} \\
\midrule
EU AI Act~\cite{raman2025navigating} & European Commission & Risk-based tiers, human oversight, transparency & High-risk system regulation, employment, health, surveillance & Regional (EU) & Legal compliance with penalties \\
NIST AI RMF~\cite{ai2023artificial} & U.S. NIST & Trustworthiness, transparency, risk mitigation & Security, privacy, robustness, explainability & Voluntary (U.S.) & Self-assessment, toolkits \\
OECD AI Principles~\cite{canton2021organisation} & OECD Nations & Human-centered values, safety, accountability & Innovation vs. risk balance, cross-border alignment & Global & Member-state adoption \\
UNESCO AI Ethics~\cite{van2023ethics} & UNESCO & Equity, inclusiveness, sustainability & Socioeconomic impact, environmental, cultural diversity & Global & Advisory with monitoring reports \\
IEEE ECPAIS~\cite{ieee2019eee} & IEEE Standards Association & Transparency, accountability, bias mitigation & Algorithmic audits, ethical design & Industry-wide & Standardization, audit checklists \\
\bottomrule
\end{tabular}
\end{adjustbox}
\end{table*}

\subsection{LLM's, VLM's and Agentic AI}
Large Language Model (LLM), Vision-Language Model (VLM)and Agentic AI have a fundamental role to play in the advancement towards AGI systems. LLM's capability of natural language understanding and VLM's which can combine visual and textual information together support the development of autonomous, adaptable and context aware AI agents that serve as the driving force for AGI. In this regard, this section discusses notable AI frameworks and models which are available currently followed by a discussion on VLMs and agentic AI as a pathway towards AGI.  One of the key techniques that enables such agentic behavior is the Tree-of-Thought reasoning framework, which equips models with the ability to explore, evaluate, and revise multiple reasoning paths. A generalized outline of this structured decision-making approach is presented in Algorithm 3.

\begin{tcolorbox}[enhanced, sharp corners, colback=gray!5!white, colframe=blue!80!black,
title=Algorithm 3: Tree-of-Thought Reasoning, fonttitle=\bfseries, boxrule=0.5pt]
\textbf{Input:} Problem description \( P \) \\
\textbf{Output:} Final solution path \( S \)

\begin{enumerate}
    \item Initialize root thought with task prompt
    \item Expand nodes with plausible reasoning paths
    \item Evaluate each path using scoring heuristics or LLM feedback
    \item Apply lookahead and backtracking to prune low-reward branches
    \item Select optimal reasoning trajectory \( S \)
\end{enumerate}
\end{tcolorbox}

\subsubsection{VLMs and Agentic AI as a pillar for the future AGI Framework}

VLMs represent a pivotal advancement in AI by integrating visual perception and linguistic understanding, enabling tasks like captioning, visual question answering, and multimodal reasoning \cite{vinyals2015show, antol2015vqa}. Rooted in early computer vision (e.g., object detection \cite{sapkotaaobject}) and NLP research (e.g., machine translation), initial approaches were constrained by their unimodal focus \cite{peters2017foundations}. The creation of paired datasets like Pascal VOC and Flickr30k \cite{pascalvoc_pwc, flickr30k_pwc} enabled learning associations between images and text. This led to the emergence of early VLMs, which combined CNN-RNN pipelines for captioning and VQA, though they often lacked deeper semantic understanding \cite{vinyals2015show}. A paradigm shift occurred with the Transformer architecture \cite{vaswani2017attention}, unifying NLP and vision through self-attention. This enabled models like BERT \cite{devlin2019bert} and ViT \cite{dosovitskiy2020image} to advance multimodal understanding, forming the backbone of contemporary VLMs increasingly applied in domains, such as robotics, medicine, and assistive technologies \cite{sapkota2025review3dobjectdetection}. 

Table~\ref{tab:agi-panels} (panel B) presents a roadmap connecting brain-inspired principles to the development of AGI via VLMs. Key \textit{brain functions} such as neocortical reasoning and hippocampal spatial memory \cite{wagner2020neocortex, vann2011hippocampus} are reflected in transformer-based architectures that employ cognitive modularity and attention mechanisms \cite{dai2023brain}, paving the way for neuro-symbolic planning \cite{sheth2024neurosymbolic} and cognitive digital twins in medical diagnostics \cite{guo2025ten}. The brain's \textit{memory hierarchies}, which transition from sensory encoding to long-term storage \cite{hannula2017beyond}, are represented in VLMs through contextual embeddings and dynamic prompt extensions \cite{dakat2024enhancing}, supporting lifelong learning and adaptive tutoring systems. In terms of \textit{action systems}, the integration of mental and physical processes \cite{schillaci2016exploration} is emulated by multi-agent VLMs and vision-action loops \cite{liu2025advances, sapkota2025concepts}. Finally, \textit{world models}-compact internal representations for prediction and planning \cite{mumuni2025large, calleo2025ai}-are realized through multimodal embeddings and simulator-based architectures, supporting anticipatory agents for household and space missions \cite{liu2025advances}. Together, these components illustrate how brain-inspired VLMs can advance AGI through the integration of embodied reasoning, hierarchical memory, and goal-directed action. 

The adoption of Transformers enabled VLMs to process images and text using unified self-attention architectures, significantly enhancing multimodal integration~\cite{bordes2024introduction}. Contrastive learning approaches, as in CLIP and ALIGN, align image-text pairs in shared embedding spaces for robust general-purpose representations~\cite{radford2021learning,jia2021scaling}. Scaling up with models like Flamingo, PaLI, and LLaVA introduced few-shot learning, multimodal dialogue, and state-of-the-art performance on diverse tasks~\cite{alayrac2022flamingo,chen2022pali,liu2023visual}.

\begin{figure*}[ht!]
  \centering
  \includegraphics[width=\textwidth]{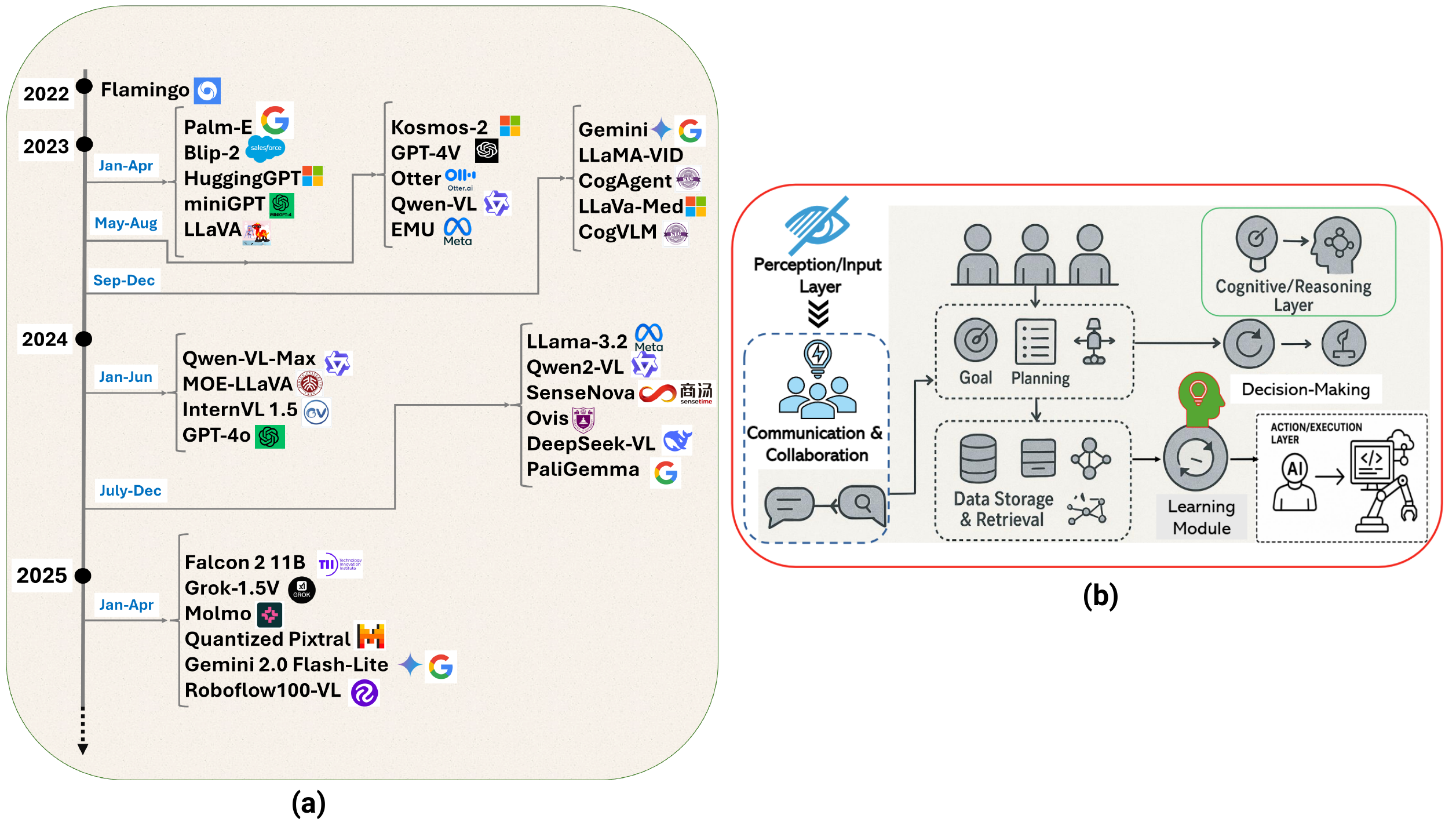}
  \caption{(a) Chronological evolution of VLMs following the release of ChatGPT in late 2022. The timeline highlights key VLM developments across major research labs and companies, organized by quarterly intervals from 2022 through early 2025. (b) Illustrating a visual overview of core functionalities in Agentic AI which is a key to AGI. This figure depicts the layered structure through which AI agents perceive inputs, make decisions, execute actions, and engage in learning and coordination to operate effectively in both individual and collaborative settings (Agentic System/MAS).}
  \label{fig:timelineEvolution}
\end{figure*}

\begin{figure*}[ht!]
  \centering
  \includegraphics[width= 0.78\textwidth]{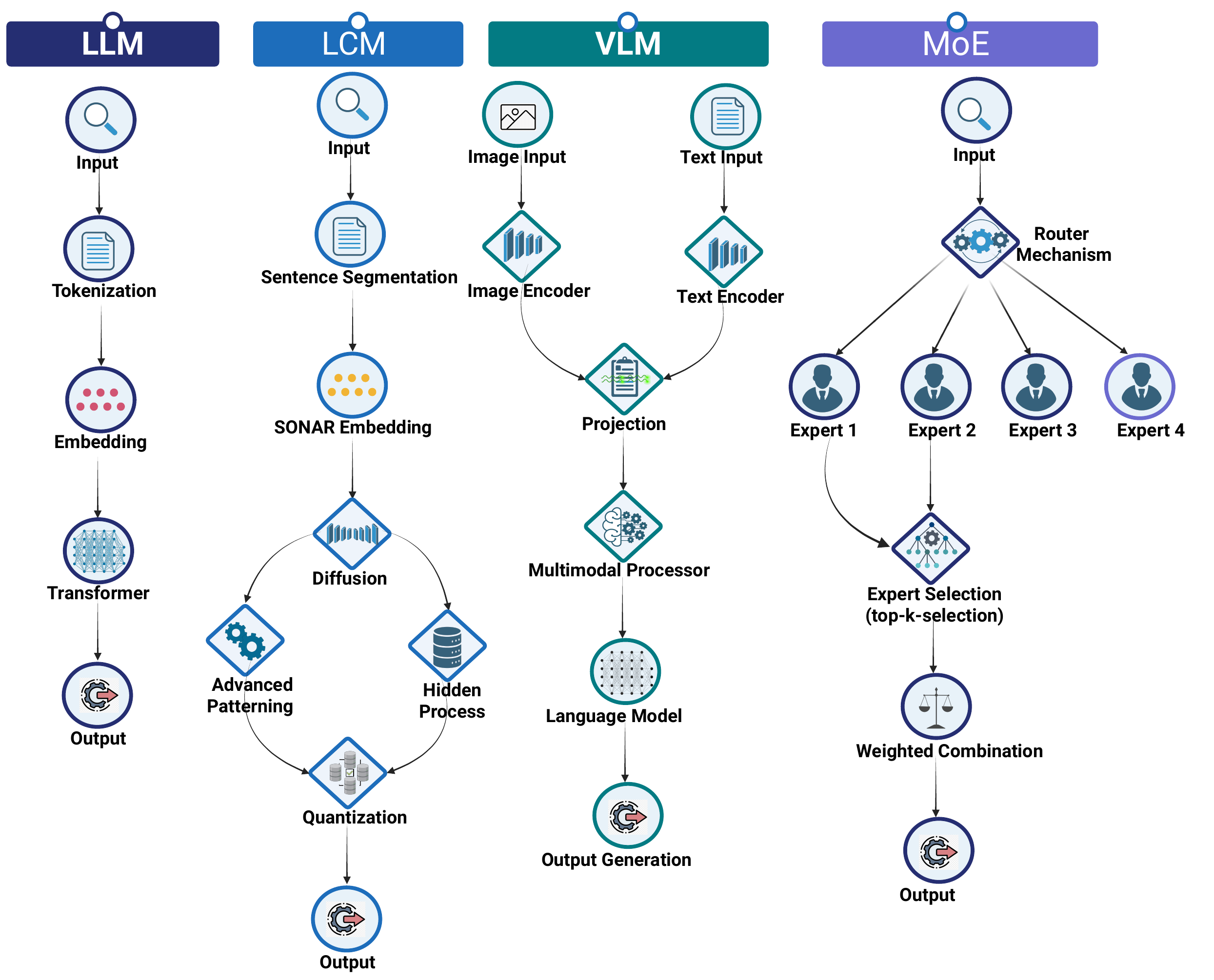}
  \caption{Conceptual overview of core foundation model architectures. The architectural pipelines of Large Language Models (LLMs), Language-Centric Models (LCMs), Vision-Language Models (VLMs), and Mixture of Experts (MoE).}
  \label{fig:concept}
\end{figure*}

Figure\ref{fig:timelineEvolution}(a) presents the chronological evolution of VLMs following the release of ChatGPT in late 2022. These models have rapidly advanced in terms of scale, multimodal comprehension, and cross-domain generalization \cite{zhang2024vision}. Current state-of-the-art VLMs support a wide spectrum of capabilities including visual question answering, captioning, visual reasoning, and image-to-text alignment. In applied domains, they have been deployed for robotic instruction following, autonomous navigation, and assistive dialogue agents. A critical advantage of VLMs lies in their ability to translate perception into semantically rich representations, enabling downstream reasoning and decision-making. Yet, despite these advances, VLMs alone cannot fulfill the requirements of AGI. They excel at perception and interpretation, but lack structured autonomy, persistent memory, and adaptive goal management. To truly transition from perception to intelligent action, VLMs must be embedded within broader Agentic AI architectures, where decision-making, coordination, and learning unfold across layered cognitive processes. 

Figure~\ref{fig:timelineEvolution}~(b) illustrates this complementary architecture. At the core of Agentic AI lies a modular framework where VLMs serve as the perceptual interface detecting objects, interpreting environments, and feeding this information into a cognitive reasoning layer. This is followed by modules for goal formulation, planning, and data storage and retrieval, which maintain contextual coherence across tasks. Agents then utilize learning modules for continuous adaptation, drawing on episodic and semantic memory to inform future actions \cite{sapkota2025vibe, acharya2025agentic}. Through collaboration and communication modules, agents interact within multi-agent systems (MAS), enabling distributed problem-solving and collective intelligence \cite{sapkota2025ai}. The decision-making layer synthesizes insights from upstream modules, and the action execution layer interfaces with external actuators or APIs to carry out commands. This layered system ensures that agent behavior is not just reactive but context-aware, goal-driven, and self-refining hallmarks of AGI. As these systems mature, Agentic AI will increasingly enable long-horizon autonomy in fields such as scientific discovery, healthcare, and adaptive robotics. By combining VLMs for perception with agentic architectures for reasoning and execution, we move closer to AGI systems that not only perceive and describe the world but also act within it with purpose, adaptability, and alignment with human values.

Additionally, the future of AGI hinges not just on increasing model scale or parameter count, but on the emergence of Agentic AI systems endowed with autonomy, memory, tool-use, and decision-making capabilities that mirror core aspects of human cognition \cite{acharya2025agentic}. Unlike static models that simply respond to prompts, Agentic AI systems act, plan, reflect, and adapt over time \cite{shavit2023practices, acharya2025agentic}. Several promising frameworks illustrate this paradigm shift: AutoGPT \cite{yang2023auto} orchestrates sequential tool calls using a planner-reflector loop; BabyAGI implements a task prioritization loop with a vector-based memory store; CAMEL (Communicative Agents for Mind Exploration of Large-scale language models) enables multiple agents to coordinate via natural language dialogue \cite{li2023camel}; ReAct fuses reasoning and acting through intermediate reasoning traces \cite{yao2023react}; and OpenAGI integrates goal-oriented decision-making with tool use and memory retrieval \cite{ge2023openagi}. Each of these systems demonstrates attributes critical to AGI, including context persistence, agent collaboration, and feedback-guided learning. When integrated with VLMs such as LLaVA \cite{liu2023visual}, Flamingo \cite{alayrac2022flamingo}, or Kosmos-2 \cite{peng2023kosmos}, these agents acquire perceptual grounding in real-world environments, enabling a more adaptive and embodied form of intelligence. 

VLMs enable agents to interpret multimodal data, including images, text, and videos, while reasoning about this information in a human-like manner \cite{sapkota2025concepts}. For example, an embodied agent equipped with VLM capabilities can interpret its environment, plan actions, and learn through interactions, mirroring how humans link perception and motor actions. This convergence is already evident in domains like robotics, assistive medical agents, and multi-agent research systems. However, a critical bottleneck persists: most current agentic systems depend on human-curated tasks, externally defined reward signals, or fine-tuned supervision, limiting their long-term autonomy and adaptability. For AGI to emerge, these agents must evolve beyond being mere tool-users; they must become self-motivated learners, capable of generating, testing, and refining their own reasoning processes. This is where the Absolute Zero paradigm presents a transformative shift.

The AZR introduces a self-evolving agentic AI paradigm that discards dependence on human-labeled tasks by autonomously generating, solving, and validating its own reasoning problems using a code execution engine \cite{zhao2025absolute}. Built on Reinforcement Learning with Verifiable Rewards (RLVR) \cite{mroueh2025reinforcement}, AZR supports outcome-based, self-verifying learning without external supervision. Its meta-cognitive curriculum design enables continuous skill refinement by identifying and addressing its own reasoning gaps. AZR is both model-agnostic and scalable, making it adaptable for integration into larger agentic ecosystems such as multi-agent research assistants or autonomous robotics. Empirically, it achieves state-of-the-art performance on mathematical and code reasoning benchmarks, outperforming traditional zero-shot models. By enabling AI systems to improve through introspective feedback rather than curated data, AZR advances AGI toward reflective, self-directed learning, pushing AI closer to human-like, adaptive, and open-ended intelligence.

In summary, future AGI will likely take the form of a self-improving, multimodal system capable of autonomous reasoning, adaptive learning, and goal-directed behavior across diverse, open-ended environments, integrating agentic AI, structured memory, and world modeling to emulate human-like cognition.

\section{Recent Advancements and Benchmark Datasets}
The pursuit of AGI has recently entered a phase defined by the emergence of increasingly general, autonomous, and multi-capable systems~\cite{moor2023foundation}. This section highlights several of the most prominent conceptual frameworks and approaches that reflect current trends in AGI design blending planning, reasoning, memory, and environmental interaction in novel ways. This is followed by a discussion on data, which is essential for AGI development. 

\subsection{Advancements Beyond Large Language Models}

\begin{figure*}[h!] % Placement specifier (h=here, t=top, b=bottom, p=page, !=force)
    \centering
    \includegraphics[width= 0.88 \textwidth]{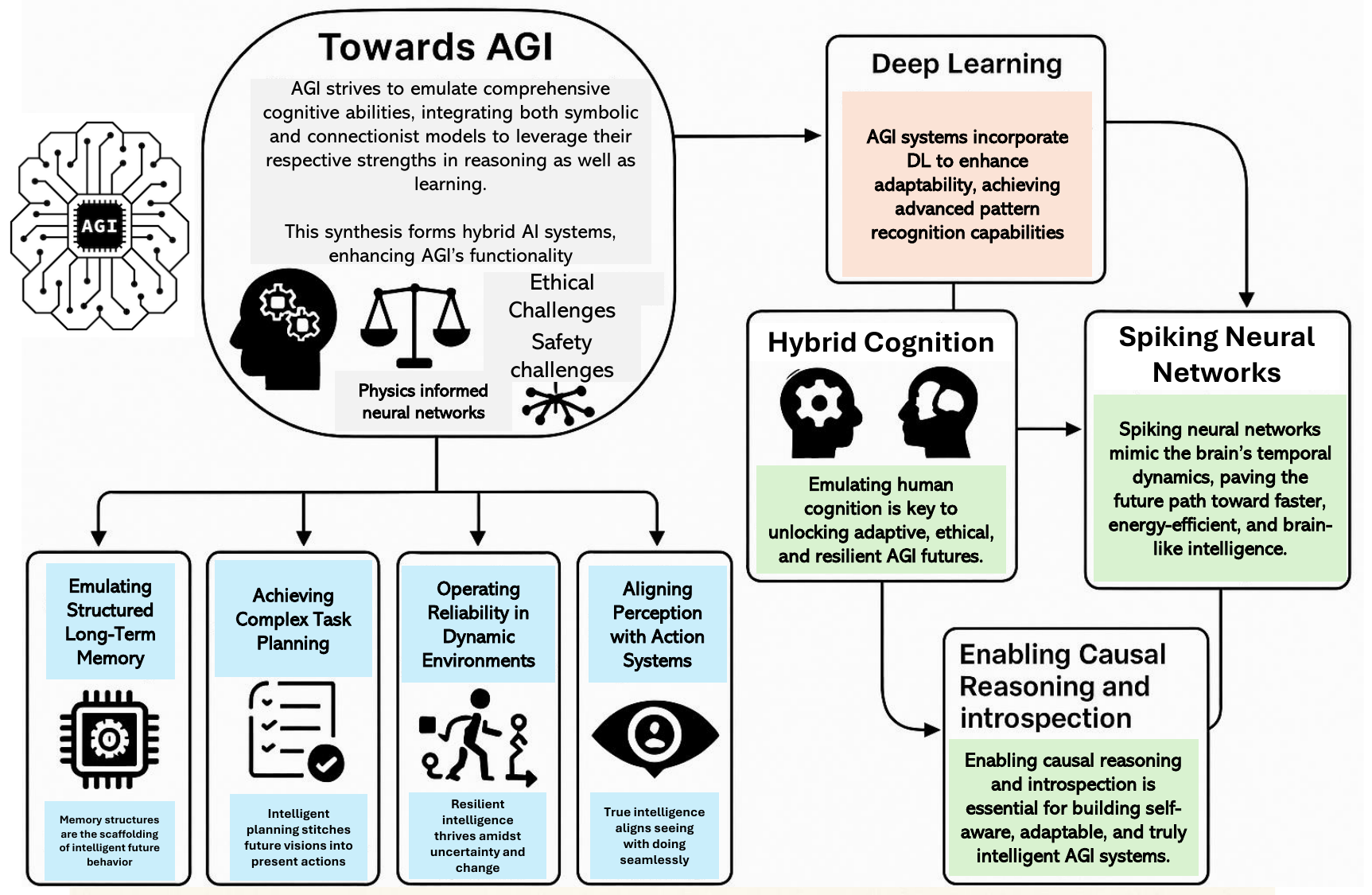} % Adjust width as needed
    \caption{Illustrating AGI’s progression toward human-like intelligence by integrating symbolic and connectionist models, emphasizing structured memory, causal reasoning, adaptive planning, and perception-action alignment, while addressing safety, efficiency, and introspective cognitive capabilities for future development. Merge our proposal}
    \label{fig:vlm_evolution2} % For referencing the figure
\end{figure*}

The progression toward AGI, as depicted in Figure \ref{fig:vlm_evolution2} necessitates overcoming the inherent limitations of current LLMs, which primarily rely on autoregressive next-token prediction. While this approach facilitates multi-task learning \cite{chang2024survey,wu2024pim}, it may not fully capture complex human cognitive processes, such as intuition and ethical reasoning \cite{mahowald2024dissociating,netz2024using}. Figure \ref{fig:intro1} illustrates AI's evolution since the 1950s, highlighting milestones where AI systems have matched or exceeded human-level performance across various domains. This historical trajectory underscores the accelerating pace of AI development, suggesting that future advancements may continue to outpace human capabilities.

The reliance on scaling laws~\cite{bahri2024explaining}, indicates that while increasing model size and training data enhances performance, this approach encounters diminishing returns~\cite{shanmugam2022learning}. Sustained scaling requires exponentially greater computational resources for increasingly marginal gains, and fundamental human abilities, such as creativity and moral reasoning, may not be effectively captured through scaling alone. This limitation underscores the need to explore more advanced learning mechanisms and architectural innovations capable of addressing the ethical and intuitive dimensions of intelligence.

\subsubsection{AI Agent Communication Protocols}

As the field advances towards AGI, robust and interpolable communication between autonomous AI agents has emerged as a critical enabler. Recent few foundational agent communication protocols such as the model context protocol (MCP) \href{https://www.anthropic.com/news/model-context-protocol}{Source Link}, the agent communication protocol (ACP) \href{https://agentprotocol.ai/}{Source Link}, the Agent2Agent protocol (A2A) \href{https://github.com/google-a2a/A2A}{Source Link}, and the agent network protocol (ANP) \href{https://agent-network-protocol.com/}{Source Link} represent key milestones in the development of scalable, compositional, and collaborative agent ecosystems.  

MCP, pioneered for LLM-centric systems such as OpenAI’s Assistants API, standardizes how models receive external tools and context through secure, typed JSON-RPC interfaces \cite{hou2025model}. This enhances context-awareness during inference and allows modular tool mounting, a cornerstone for generalizable intelligence. ACP further advances this by enabling REST-native, session-aware messaging between heterogeneous agents with structured MIME-typed payloads, fostering reliable multimodal coordination. A2A introduces a peer-to-peer framework where agents advertise capabilities via dynamic “Agent Cards” and negotiate task delegation through structured artifacts \href{https://github.com/google-a2a/A2A}{Source Link}. This supports fine-grained collaboration between agents across frameworks and vendors, promoting agent autonomy and specialization. Likewise, ANP pushes the frontier with decentralized, internet-scale discovery and collaboration, using DID-authenticated agents and semantic web standards (JSON-LD, Schema.org). It establishes the foundation for federated agent networks with open trust and runtime negotiation.

Together, these protocols define a layered infrastructure for communication, identity, and task management. They collectively support the emergence of agent societies capable of distributed reasoning, adaptive coordination, and persistent memory \cite{sapkota2025ai, shamsujjoha2025swiss, jabbour2024generative}, hallmark of the AGI systems. Their evolution marks a shift from isolated, monolithic agents toward scalable, interoperable networks of intelligent entities operating with shared context and collective goals.

\subsubsection{Large Concept Models}
As AI technology advances towards AGI, the underlying bottlenecks of token-level processing have become increasingly apparent, driving the development of architectures that operate at higher level of semantic abstraction \cite{barrault2024large}. Large Concept Models (LCMs) are a quantum leap from token-level language prediction models to concept-level reasoning-based language prediction models (Figure \ref{fig:concept}), providing the machine with a human-like manner of understanding and processing language, which is consistent with hierarchical cognitive process. 

LCMs are designed to operate over explicit higher-level semantic representations known as “concepts”, which are language- and modality-agnostic abstractions that represent ideas or actions in a structured flow.  Unlike LLMs, which process the text at token level, LCMs predict the next concept rather than the next token, with every concept being a sentence-level semantic representation. This architectural novelty is enabled possible by the SONAR embedding space \cite{duquenne2023sonar}, a multilingual and multimodal fixed-size sentence embedding framework that supports more than 200 languages in text and 76 languages in speech and supports the concept-level reasoning through its intricate encoder-decoder model.

LCMs are a critical building block in the pursuit of AGI, as they enable AI systems to work in terms of concepts rather than individual words, thereby allow for the development of deep contextual understanding and more coherent long-form generation. The development of LCMs represents a fundamental paradigm shift from token-based language modeling towards a semantic-based language modeling, offering a closer approximation of human cognitive processes without the limitation imposed by modality competition \cite{aghajanyan2022cm3}

\subsubsection{Large Reasoning Models (LRMs)} 
% Recent analyses of Large Reasoning Models (LRMs) indicate that while models exhibit sophisticated reasoning traces, their performance drastically declines when faced with increasingly complex tasks. This underscores a fundamental limitation: their inability to generalize effectively under uncertainty, especially beyond a certain threshold of task complexity~\cite{illusion-of-thinking}. 
LRMs represents a shift away from traditional language models, moving toward systems that focus on explicit, multi-step cognitive processes as opposed to single-shot response generation \cite{wei2022chain}. This method derives from human problem-solving behavior, in which complicated problems are analyzed in sequences of the reasoning process, nested on previous conclusions. Extended inference time computation lies at the core of LRMs and involves the training of models to ‘think’ through problems in a structured manner, as opposed to relying only on pattern matching from already seen training examples \cite{wu2024comparative}. These systems employ techniques including chain-of-thought reasoning, self-reflection, and iterative refinement to generate more accurate and well-reasoned outputs \cite{shinn2023reflexion}. This controlled computational approach allows models to perform advanced mathematical, logical, and analytic operations, far exceeding the capabilities of even the largest autoregressive language models.

The LRM paradigm changes the typical trade-off between model size, computational complexity and performance by showing that the computation resources can be spent effectively on the inference side rather than the training side \cite{cobbe2021training}. Unlike typical architectures which learn responses in a single forward pass, LRMs perform prolonged reasoning processes, and sometimes require multiple iterations, self-correction, and fact-checking. This mirrors human cognition, for which hard problems require attention, working memory, and systematic cycling through possible solution paths before a non-intuitive solution occurs.

The reasoning-centered design of LRMs mirrors the structured nature of human reasoning during analytical thought, where complex problems are approached via effortful decomposition, hypothesis generation and evidence scrutiny. This systematic treatment of problems is key to the development of more robust and interpretable AI systems that deal with tasks that start from real understanding of the data, instead of merely patterns that arise from the data.

\subsubsection{Mixture of Experts}
Mixture of Experts (MoE) is a departure from monolithic neural network architectures, considering models as ensembles of specific sub-networks, selectively triggered by the input \cite{ cichocki2021future}. This argument is based on the biological analogy of modular architecture, typical of some parts of the brain specializing in processing different kinds of information \cite{ papi2021mixtures}. At the center of MoE are multiple “expert” networks, each of which can handle part of the overall task, and a “gating” network that dynamically chooses to which experts to send its inputs \cite{fedus2022switch}. Such conditional computation enables a much higher model capability to be achieved without a linear increase in computational cost. The gating mechanism is learned to distribute the computation across the experts, such a way that only a small fraction of parameters are activated for each given input \cite{gan2025mixture}. This is in contrast to traditional dense neural networks, where all parameters need to participate in processing each sample, resulting in huge computational cost as the model grows \cite{sun2025invmoe}.

The MoE paradigm, which promotes a specialized yet coordinated intelligence architecture, mirrors human cognition where the brain consists of specialized physical regions which are specialized in different functions, yet capable of seamlessly integrating to solve complex tasks. It is widely believed that this modularity and specialization are essential for the efficiency, adaptability, and plasticity of human intelligence.

\subsubsection{Neural Society of Agents}
Another approach towards decentralized decision making and prediction is Neural Society of Agents. Within this, rather than a single model that is all encompassing, the neural society of agents approach suggests a multi-agent AI model, in which different agents have distinct expertise and that share intelligence to collaborate on solving complex problems \cite{Korteling2021}. This resembles the system, found in nature, in which individual cells or organisms work together to achieve overall goal \cite{cichocki2021future}. This method also supports distributed problem decomposition and task assignment, since capabilities are distributed amongst the agents, leading to a parallel implementation and enhanced efficiency. Moreover, the interactions between agents can lead to an enhanced collective intelligence which can be greater than that of any single agent, such as found in social insect's colonies \cite{tallam2025autonomous}.
To achieve the above functionality, the neural society of agents requires work in multiple areas such as multi-agent reinforcement learning, optimizing communication protocols, coordination mechanisms and managing emergent behaviors \cite{de2025open}.

The creation of neural societies of agents represents a compelling approach to AGI, as it reflects the distributed and collaborative nature of human intelligence. Human cognition is not a unitary construct, but rather the product of complex interactions among multiple cognitive modules and brain regions. By developing communities of artificial agents that can collaborate, share their findings and learn from each other, we may be able to replicate some of the most powerful attributes of human intelligence and ultimately enabling the creation of more general, adaptive and flexible AGI systems. 

\subsection{The importance of benchmark datasets} 
Benchmark datasets have been foundational to progress in AI, enabling fair comparisons and standardizing evaluations, e.g., ImageNet for vision~\cite{deng2009imagenet}, GLUE, HELM, and ALM-Benc for language~\cite{wang2018glue, liang2022holistic, vayani2025all}. However, current benchmarks often assess narrow capabilities and fall short of testing generalization, long-horizon planning, or socio-cognitive reasoning key to AGI. To evaluate AGI systems meaningfully, we need next-generation benchmarks that integrate multi-modal inputs, real-world constraints, ethical reasoning, and interactive environments. Initiatives like ARC~\cite{chollet2019measure} and BIG-Bench~\cite{srivastava2022beyond} point in this direction, but broader, dynamic benchmarks are still lacking. Table~\ref{tab:agi-panels} summarizes the prominent benchmarks used to evaluate the capabilities related to AGI in reasoning, embodiment, and language interaction.

\subsection{The Role of Synthetic Data in AGI} 

Synthetic data has emerged as a pivotal component in scaling and generalizing AI systems, offering controllable diversity, infinite augmentation, and safe simulation for high-risk or rare scenarios~\cite{jordon2022synthetic}. Procedurally generated environments such as BabyAI and MineDojo~\cite{fan2022minedojo} enable agents to train in highly customizable tasks, while self-play and emergent curricula exemplified by AlphaZero and Voyager allow for autonomous skill acquisition without explicit supervision~\cite{bertsekas2022lessons}. 
\par Moreover, LLMs now routinely generate synthetic instruction–response datasets, accelerating pretraining and fine-tuning pipelines. However, the misuse of synthetic data can lead to systemic biases, factual drift, and ethical misalignment, especially when artificial distributions diverge from real-world human contexts~\cite{anderljung2024protecting}. As AGI systems grow more autonomous and capable, ensuring the quality, representativeness, and traceability of synthetic data has become essential for developing robust, grounded, and ethically aligned intelligence~\cite{raza2024unlocking}.

\section{Missing Pieces and Avenues of Future Work}
 
While there has been enormous progress towards the goal of AGI, there are several aspects that still are missing. A major issue with current systems in terms of AGI is the lack of true creativity and innovation. Currently available models excel at using already seen data to generate outputs, they still lack true creativity capability. AGI systems need to be able to "think out of the box" which requires pushing the boundaries posed by the confines of input data. 
\subsection{Uncertainty in AGI: Navigating a Dual-Natured Universe}

AGI aspires to emulate human-like intellectual versatility, crucially including managing uncertainty inherent in our dual-natured universe, where deterministic rules coexist with random, unpredictable events \cite{Korteling2021, gupta2025personalized}. Unlike narrow AI, optimized for structured environments, AGI must autonomously adapt and make informed decisions under conditions of incomplete knowledge and inherent randomness.

Two principal uncertainty types confront AGI. \textit{Epistemic uncertainty}, reflecting deterministic limitations, arises from incomplete or noisy data, training gaps, or novel environments beyond prior knowledge \cite{Korteling2021}. In contrast, \textit{aleatory uncertainty} captures the intrinsic randomness of natural and social phenomena, such as unpredictable human emotions or environmental variability that defy deterministic modeling regardless of data quantity \cite{Kuznietsov2024, Hoel2023}.

Effectively navigating these uncertainties requires AGI to dynamically balance exploration of new knowledge and exploitation of established information, thereby enabling optimal decision-making in unpredictable settings \cite{stanovsky2025beyond, saeed2023explainable}. Additionally, decisions under uncertainty carry profound ethical implications, necessitating interpretable and accountable AGI systems to mitigate biases, unfair outcomes, and unintended consequences \cite{guan2022ethical, hassija2024interpreting}.

\begin{tcolorbox}[colback=mybluebg, colframe=myclueborder, colbacktitle=mybluetitle, coltitle=black, fonttitle=\bfseries, top=2pt, bottom=2pt, title=The Dual Universe: Random and Deterministic Dynamics in AGI]
While the universe is inherently stochastic, AGI systems equipped with continual learning mature by absorbing real-world variance. Over time, uncertainty becomes compressible into structured knowledge facilitating robust, deterministic adaptation and generalization.
\end{tcolorbox}
\subsection{Beyond Memorization: Compression as a Bridge to Reasoning}
The success of Large AI systems much still stems from memorization at scale, since these models are trained to predict the next token, these models often fails in unfamiliar situations~\cite{morris2025much}; particularly those demanding causal reasoning~\cite{cau2023effects} long-horizon planning~\cite{fengfar}, or physical intuition~\cite{hadi2023large}. 
%This raises a fundamental question: how do we transition from memorizing to truly understanding?

\paragraph{Reasoning and Memorization Are Not Opposites}

Reasoning and memorization are considered distinct or even opposing capabilities~\cite{xie2024memorization}. In reality, they exist on a continuum shaped by the degree to which information is compressed~\cite{kyllonen1990reasoning}. Memorization corresponds to low compression, which means that one simply stores examples like a lookup table. True reasoning reflects high compression, abstracting core principles and applying them flexibly to novel problems!\cite{hu2023chatdb}.

Most LLMs operate between these extremes. They don’t merely memorize—they generalize shallowly by interpolating across known patterns. Yet this is not full abstraction. Their reasoning remains fragile, limited by training data and lacking mechanisms for grounding or principled inference~\cite{li2024llms}.

\paragraph{Designing for Compression and Abstraction in AGI}

The path forward isn’t to discard memory, but to structure it more intelligently. Memorization supplies facts; reasoning turns those facts into insights. AGI will require architectures that embrace both—using tools like retrieval-augmented generation (RAG)~\cite{zhang2024interactive}, modular reasoning agents~\cite{costantini2021epistemic}, and memory-aware training strategies that encourage deeper compression~\cite{tishby2000information}.

\begin{tcolorbox}[colframe=myclueborder, colbacktitle=mybluetitle, coltitle=black, fonttitle=\bfseries, top=2pt, bottom=2pt, title=Decomposing Intelligence: Reasoning + Memory]
While memory and reasoning are often seen as separate, true intelligence arises from their synergy. Memory anchors past experience; reasoning abstracts and applies it to new situations. Their integration enables adaptive, context-aware behavior—central to AGI design.
\end{tcolorbox}

%True general intelligence depends not just on size, but on the system’s ability to distill experience into portable, reusable knowledge. This is the shift from surface fluency to meaningful understanding—from vast recall to real thinking.

\subsection{Emotional and Social Understanding}

Current AI systems lack the capacity  to perceive emotions or navigate complex social dynamics. For AGI to achieve human-level intelligence, it must engage with users in emotionally, empathatically and context-aware ways~\cite{scribano2023ai}. This requires integrating psychological theories , human behavioral data, and leveraging multimodal learning techniques to effectively detect, interpret, and respond to emotional and social cues effectively.
\subsubsection{Ethics and Moral Judgement}
True AGI must operate within a comprehensive ethical and moral framework. Event current systems, despite lackin general intelligence, exhibit biases that raise concerns~\cite{li2025should}. To prevent harmful outcomes, AGI development must embed ethical principles from the outset, guided by interdisciplinary consensus among legal, ethical, and sociological experts. Furthermore, AGI systems should incorporate human-in-the-loop feedback mechanisms to ensure accountability and promote responsible behavior~\cite{boltuc2022moral}.

\subsection{Debt in the Age of AGI: Cognitive and Technical Risks}
One emerging concern is \textbf{cognitive debt}, a long-term erosion of human intellectual engagement caused by overreliance on LLMs. Recent neurobehavioral studies~\cite{kosmyna2025your} reveal that participants using LLMs exhibit reduced neural connectivity, lower recall, and diminished essay ownership compared to those relying on their own cognition. %These effects intensify over time, as users gradually offload reasoning, planning, and synthesis, trading convenience for cognitive disengagement.

\paragraph{Technical Debt}In parallel, AGI development is accelerating the phenomenon of \textbf{technical debt} through practices like \textit{vibe coding}~\cite{chow2025technology}, where code is generated based on surface-level pattern completion rather than robust logic or modular design.

These dual debts, whether cognitive and technical, are not peripheral concerns. They reflect a broader imbalance in current AGI trajectories: prioritizing short-term performance and usability over foundational understanding and resilience~\cite{boltuc2024human}. Mitigating them requires not only architectural guardrails, but also thoughtful co-evolution of education, software engineering norms, and human-AI interaction design.

\subsection{Power Consumption and Environmental Impact}

The infrastructure supporting computationally intensive models demands immense electricity, with projections indicating substantial increases as development advances toward AGI~\cite{IEA_AIelectricity}. This escalating energy consumption not only limits scalability but also exacerbates environmental concerns, including carbon emissions and resource depletion. To mitigate these impacts, AGI development must prioritize energy-efficient model architectures, low-power deployment strategies, and sustainable data center operations~\cite{singh2024dynamic}. 
% Mitigating the environmental footprint of AGI is essential for responsible and scalable adoption.
\section{Our Proposal} We propose that true general intelligence agent can engineered by decomposing the general intelligence into several componensts, such as spatial intelligence, geometrical intelligence, sequential intelligence, multi-modal intelligence, and social intelligence, with each component engineered both in isolation and as holistic, under dynamic real-world scenarios.

\section{Conclusion}\label{conclusion}
AGI remains one of the most profound scientific challenges of our time, demanding not only greater scale, but also deeper alignment with the cognitive, ethical, and societal foundation of human intelligence. This paper has examined AGI from a multidisciplinary lens, synthesizing insights from neuroscience, symbolic reasoning, learning theory, and social systems design. We argue that current paradigms, especially those grounded in next-token prediction are insufficient to yield agents capable of robust reasoning, self-reflection, and generalization across unstructured, uncertain environments. 

Several challenges remains, such as the need for grounded world models, dynamic memory, causal reasoning, robust handling of aleatory and epistemic uncertainty, developing perception of emotional and social contexts and collective agent architectures. Significant advancements have been made, such as Large Concept Models, Large Reasoning Models and Mixture of Experts, which improve LLM performance beyond next-token prediction by incorporating biologically inspired behaviors into output generation. The  "society of agents" metaphor offers a promising direction, reflecting both biological modularity and the need for specialization and internal negotiation in future AGI systems.

Looking forward, we believe that true progress toward AGI will require a fundamental shift from monolithic models to modular, self-adaptive, and value-aligned systems. This transition must be accompanied by social foresight, involving the proactive redesign of education, labor, and policy frameworks to accommodate and co-evolve with intelligent machines. AGI cannot be purely a technical pursuit. On the contrary, it must be a human project with development progressing alongside humans actively involved in the process. This requires the inclusion of diverse stakeholders in the development process through cultivating a shared, inclusive vision and goal-setting. Such an ecosystem will facilitate the responsible and socially acceptable advancement of AGI.

\begingroup
\small 
\balance
\bibliographystyle{unsrt}  

\bibliography{sample-base}
\endgroup
\newpage
% \begin{document}
%\title{Supplementary}
\renewcommand{\thefigure}{A\arabic{figure}} 
\renewcommand{\thetable}{A\arabic{table}} 
\renewcommand{\theequation}{A\arabic{equation}} 

\setcounter{figure}{0}
\setcounter{table}{0}
\setcounter{equation}{0}

\onecolumn

\section*{Appendix}
\scriptsize
%\textbf{Glossary of Terms}
\setlength{\LTleft}{0pt}
\setlength{\LTright}{0pt}
\begin{longtable}{p{5cm} p{2cm} p{8cm}}
% \centering

\caption{Glossary of Terms}
\label{tab:glossary}\\
\toprule
\textbf{Term} & \textbf{Abbreviation} & \textbf{Definition} \\
\midrule
\endfirsthead

\multicolumn{3}{c}{} \\
\toprule
\textbf{Term} & \textbf{Abbreviation} & \textbf{Definition} \\
\midrule
\endhead

\midrule
\multicolumn{3}{r}{} \\
\endfoot

\bottomrule
\endlastfoot

Abstract Reasoning Corpus & ARC & Benchmark that evaluates abstract reasoning and pattern-completion skills beyond surface pattern matching. \\
Agent Communication Protocol & ACP & Communication system designed for software agents allowing them to communicate using RESTful protocol. \\
Agent Network Protocol & ANP & Decentralised protocol using decentralized identifiers and semantic-web standards for discovery and collaboration among federated agents. \\
Agent2Agent Protocol & A2A & Peer-to-peer protocol where agents advertise capabilities via agent cards and negotiate task delegation. \\
ALIGN & ALIGN & Google vision–language model trained on noisy web-scale image–alt-text pairs for universal cross-modal representations. \\
AlphaFold2 & AlphaFold2 & Google DeepMind's AI system that predicts protein structure from amino acid sequences with high accuracy, revolutionizing structural biology. \\
AlphaGo & AlphaGo & Google DeepMind's reinforcement learning system that defeated world champions in the game of Go, combining deep neural networks with Monte Carlo tree search. \\
Application Programming Interface & APIs & Standardised interfaces that let separate software components communicate and exchange functionality or data. \\
Abstract Reasoning Corpus & ARC & Visual reasoning benchmark created by Francois Chollet that consists of puzzles where you need to figure out the underlying pattern or rule. \\
Artificial General Intelligence & AGI & Systems capable of flexible, human-level reasoning and learning across domains, without task-specific retraining. \\
Automated Language Model & ALM & Systematic approach to evaluating language models using automated testing procedures across multiple benchmarks and tasks without manual intervention.\\
AutoGPT & AutoGPT & Open-source agent that plans subtasks and calls tools autonomously via a planner–reflector loop over an LLM. \\
BabyAGI & BabyAGI & Minimal task-execution loop that prioritises tasks and stores context in a vector memory, driven by an LLM. \\
Beyond the Imitation Game Benchmark & BIG-Bench & Collaborative benchmark featuring diverse, challenging tasks designed to test capabilities beyond current language model performance. \\
CAMEL & CAMEL & Framework where two role-playing LLM agents collaborate via natural-language dialogue to solve tasks. \\
Cerebellum & Cerebellum & Brain region responsible for motor control, balance, and coordination, also involved in cognitive functions like language and learning. \\
Chain-of-Thought Prompting & CoT & A prompting technique that decomposes complex reasoning into interpretable sub-steps, improving performance on multi-step tasks. \\
CICERO & CICERO & Meta AI agent that achieved human-level performance in the game Diplomacy via strategic planning and natural-language negotiation. \\
Cognitive Debt & CD& Prolong reliance on AI may cause a gradual erosion of neural engagement, memory consolidation, and critical reasoning \\
Communicative Agents for Mind Exploration of Large Language Models & CAMEL & Framework enabling multiple role-playing LLM agents to collaborate via natural-language dialogue to solve complex tasks. \\
Computational Intelligence & CI & Umbrella field covering neural, evolutionary, fuzzy and swarm methods aimed at adaptive, intelligent behaviour. \\
Contrastive Language–Image Pre-training Model & CLIP & Contrastive Language–Image Pre-training model aligning textual and visual embeddings for zero-shot recognition. \\
Convolutional Neural Networks & CNNs & Neural network architectures that apply convolutional filters to capture spatial hierarchies in image data (e.g., edges → textures → objects). \\
Decentralized Identifier & DID & W3C standard for verifiable, self-sovereign digital identities that enable secure, decentralized authentication and authorization. \\
Deep Learning & DL & Sub-field of machine learning that trains deep (multi-layer) neural networks to learn hierarchical feature representations. \\
Deep Q-Network & DQN & Deep reinforcement learning algorithm that combines Q-learning with deep neural networks to learn optimal actions in complex environments. \\
Direct Preference Optimization & DPO & An alignment technique that trains models directly from human preference data, effectively turning an LLM into its own reward model for improved alignment. \\
Dual Memory Network & DMN & Architecture maintaining separate memory systems for different types of information, enabling flexible retrieval and reasoning. \\
Electroencephalography & EEG & Non-invasive neuro-imaging technique that records electrical activity via scalp electrodes, giving millisecond-level temporal resolution. \\
Electrocorticography & ECoG & Invasive recording of cortical surface potentials, offering higher spatial fidelity than EEG for research or clinical use. \\
ELIZA & ELIZA & Early chatbot developed in the 1960s that simulated conversation by using pattern matching and substitution methodology. \\
Episodic Memory & EM & The ability to recall and reuse specific past experiences, enabling context-aware reasoning and learning from interactions over time.\\
Explainable AI & XAI & A domain focused on making AI systems transparent and interpretable, embedding interpretability through neuro-symbolic reasoning, causal modeling, or attention mechanisms. \\
Flamingo & Flamingo & DeepMind vision-language model that performs few-shot image+text tasks via contrastive pre-training and frozen LLM backbone. \\
Frontoparietal Network & FPN & Large-scale brain network linking frontal and parietal cortices, implicated in executive control, attention, and flexible cognition. \\
Functional Magnetic Resonance Imaging & fMRI & Measures brain activity indirectly via blood-oxygen (BOLD) signals, producing whole-brain maps with millimetre spatial resolution. \\
General Language Understanding Evaluation & GLUE & Benchmark suite for evaluating natural language understanding across multiple tasks including sentiment analysis and textual entailment. \\
Gradient-weighted Class Activation Mapping & Grad-CAM & Explainability technique that produces visual explanations for CNN predictions by highlighting important regions in input images. \\
Group Relative Policy Optimization & GRPO & A method that optimizes reasoning quality by comparing multiple generated trajectories, improving alignment through relative policy evaluation. \\
Hippocampal & Hippocampal & Relating to or involving the hippocampus brain region, particularly in context of memory formation and spatial processing capabilities. \\
Hippocampus & Hippocampus & Brain region crucial for memory formation, spatial navigation, and learning, serving as a key inspiration for AI memory architectures. \\
Holistic Evaluation of Language Models & HELM & Comprehensive framework for evaluating language models across accuracy, calibration, robustness, fairness, bias, and efficiency. \\
Implicit Regularization & IR & Phenomenon where optimization methods (like SGD) naturally bias models toward solutions with better generalization properties. \\
Information Bottleneck & IB & A theoretical framework positing that models generalize well by compressing inputs into compact latent representations that retain only task-relevant information. \\
JavaScript Object Notation for Linked Data & JSON-LD & Method of encoding linked data using JSON, enabling semantic web standards and structured data representation. \\
\\JavaScript Object Notation Remote Procedure Call & JSON-RPC & Lightweight remote procedure call protocol using JSON for data interchange, enabling standardized communication between systems. \\
Kolmogorov–Arnold Networks & KANs & Networks using learnable spline-based activation functions rather than fixed ones, improving interpretability and flexibility in approximating complex functions. \\
Kullback-Leibler Divergence & KL & Measure of difference between probability distributions, commonly used in variational inference and information theory. \\
Large Action Models & LAMs & Foundation models that predict full action sequences (such as  API calls, tool invocations) rather than next-word tokens, enabling embodied or tool-augmented decision making. \\
Large Language Models & LLMs & Large-scale models trained on massive text corpora for language understanding and generation. \\
Large Reasoning Models & LRMs & AI systems focusing on explicit, multi-step cognitive processes and extended inference-time computation for enhanced reasoning capabilities. \\
Learning to Think & L2T & Meta-learning paradigm where an agent improves its own reasoning procedure, not just task performance. \\
LeNet-5 & LeNet-5 & Convolutional neural network architecture developed by Yann LeCun for handwritten digit recognition. \\
Locked-image Tuning & LiT & Vision-language model focusing on efficient image-text alignment and generative capabilities for multimodal tasks. \\
Low-Rank Adaptation & LoRA & Parameter-efficient fine-tuning method that adapts large models by learning low-rank decompositions of weight updates. \\
Magnetoencephalography & MEG & Neuro-imaging that detects magnetic fields generated by neuronal currents, allowing source-localised brain-activity mapping. \\
Masked Autoencoder & MAE & Vision model pre-trained by reconstructing masked image patches, yielding strong features for downstream tasks. \\
MineDojo & MineDojo & Framework for open-ended agent learning in Minecraft, providing diverse tasks and environments for embodied AI research and evaluation. \\
Minimum Description Length & MDL & A principle from algorithmic information theory stating that the simplest model that best compresses the data will generalize more effectively. \\
Mixture of Experts & MoE & Neural architecture using a gating network to route each input to a small subset of specialised expert subnetworks. \\
Model Context Protocol & MCP & Specification for passing shared context (goals, world state) among heterogeneous models/agents in a pipeline. \\
Model-Agnostic Meta-Learning & MAML & Meta-learning algorithm that finds parameter initializations enabling fast adaptation to new tasks with minimal gradient steps. \\
Momentum Contrast & MoCo & Contrastive learning approach using a momentum-updated encoder to maintain consistent representations across training batches. \\
Multi-Agent Systems & MAS & Systems composed of multiple interacting agents that coordinate to perform complex tasks via communication and shared goals. \\
Multi-Layer Perceptrons & MLPs & Feedforward neural networks with multiple hidden layers, capable of learning complex nonlinear mappings between inputs and outputs. \\
Multipurpose Internet Mail Extensions & MIME & Standard defining format of email messages and, by extension, format of content in web communications and API interactions. \\
MYCIN & MYCIN & Early expert system developed in the 1970s for diagnosing bacterial infections and recommending antibiotics, representing rule-based AI approaches. \\
National Institute of Standards and Technology & NIST & U.S. federal agency developing technology standards, including frameworks for AI risk management and trustworthiness. \\
Natural Language–based Society of Mind & NLSOM & A modular architecture composed of multiple specialized agents that communicate via natural language, enabling collaborative reasoning and problem solving. \\
Neocortex & Neocortex & The outer layer of the cerebral cortex in mammals, responsible for higher-order cognitive functions including sensory perception, motor commands, and abstract reasoning. \\
Neural Tangent Kernel & NTK & A perspective showing that infinitely wide neural networks behave like kernel regressors during training, characterizing regimes of robust generalization. \\
NIST AI Risk Management Framework &NIST AI RMF& Framework promoting AI trustworthiness through interpretability, risk mitigation, security, privacy, and robustness guidelines. \\
\\Not Safe for Work & NSFW & Content classification system used to identify material inappropriate for professional or public settings, important for AI safety. \\
Occipital Lobes & Occipital Lobes & Brain regions primarily responsible for visual processing, containing the primary visual cortex and associated visual areas. \\
Organisation for Economic Co-operation and Development & OECD & International organization developing economic and social policy guidelines, including principles for AI governance. \\
PAC-Bayes Bounds & PAC-Bayes & Theoretical framework that upper-bounds generalisation error using a prior/posterior KL-divergence term. \\
Parameter-Efficient Fine-Tuning & PEFT & Techniques (such as LoRA, adapters) that adapt a large model by only training a small subset of parameters. \\
Parietal Lobes & Parietal Lobes & Brain regions involved in spatial processing, attention, and sensorimotor integration, crucial for coordinating perception and action. \\
Partial Differential Equations & PDEs & Mathematical equations describing relationships between functions and their partial derivatives, often encoding physical laws in PINNs. \\
Pascal Visual Object Classes & Pascal VOC & Benchmark dataset for object detection and image segmentation, instrumental in advancing computer vision research. \\
Pathways Language and Image Model & PaLI & Google's multilingual, multimodal model combining visual and textual pre-training for cross-modal understanding. \\
Physics-Informed Neural Networks & PINNs & Models that incorporate physical laws (such as partial differential equations) into their architecture, ensuring predictions remain consistent with known physics. \\
Positron Emission Tomography & PET & Imaging that uses radiotracers to capture metabolic or molecular processes, often combined with CT/MRI for anatomy. \\
Proximal Policy Optimization & PPO & An RL algorithm that balances policy improvement with stability by constraining updates to a trust region in policy space. \\
Q-Learning & Q-Learning & Model-free reinforcement learning algorithm that learns optimal action-value functions through temporal difference updates. \\
ReAct & ReAct & Prompting strategy that interleaves reasoning traces and actions, letting an LLM decide when to think or call a tool. \\
Recurrent Neural Networks & RNNs & Neural network architectures designed for sequential data, maintaining hidden states to capture temporal dependencies (such as time series, language). \\
Reinforcement Learning & RL & A learning paradigm where agents learn by interacting with the environment through trial-and-error to maximize cumulative reward. \\
Reinforcement Learning with Human Feedback & RLHF & A method that incorporates human judgments into the reinforcement learning reward loop to improve alignment and safety of learned behaviors. \\
Retrieval-Augmented Generation & RAG & A technique that augments model outputs by retrieving relevant external documents or knowledge during inference, improving factual accuracy. \\
Retrieval-Enhanced Transformer & RETRO & Architecture augmenting language models with retrieval mechanisms to access external knowledge during generation. \\
Self-Evolving Agentic AI & AZR & Research project exploring agents that autonomously update their policies, memories and objectives over long horizons. \\
Sentence-level Multimodal and Language-Agnostic Representations & SONAR & Multilingual, multimodal embedding framework supporting 200+ languages for cross-lingual and cross-modal understanding tasks. \\
Simple Contrastive Learning of Representations & SimCLR & Self-supervised learning method that learns representations by maximizing agreement between differently augmented views of data. \\
Small Language Model & SLM & Compact LLM (approx.100 M–1 B parameters) optimised for edge devices or cost-sensitive deployment. \\
Spike-Timing-Dependent Plasticity & STDP & Neurobiological learning rule where synaptic strength changes based on precise timing of pre- and post-synaptic neural spikes. \\
Spiking Neural Networks & SNNs & Biologically inspired networks that emulate neural spike dynamics (such as synaptic plasticity, spike timing), enabling event-driven, energy-efficient temporal processing. \\
Stochastic Gradient Descent & SGD & First-order optimisation algorithm that updates parameters using mini-batch estimates of the gradient. \\
Structural Equation Models & SEMs & Statistical models encoding causal relationships between variables, used in causal inference and representation learning. \\
Synaptic Activities & Synaptic & Electrochemical processes at neural connections that transmit information between neurons, including excitatory and inhibitory signals essential for all cognitive functions. \\
Temporal Lobes & Temporal Lobes & Brain regions housing auditory processing areas, memory structures (including hippocampus), and language comprehension areas. \\
Test-Time Adaptation & TTA & Techniques enabling models to adapt at inference time to distributional shifts, either by optimizing certain parameters on the test batch (optimization-based) or by modifying inference behavior without weight updates (training-free). \\
Test-Time Prompt Tuning & TPT & Lightweight variant of TTT that updates only soft prompts or prefix tokens at inference time. \\
Test-Time Training & TTT & Adapts a model on the test batch itself (usually self-supervised) to counter distribution shift during inference. \\
Trajectory Modelling & Trajectory Modelling & Framework that treats multi-step decision sequences as fundamental units for modeling, enabling AI systems to plan over extended horizons. \\
Training-Free Dynamic Adapter & TDA & Test-time adaptation approach that modifies inference behavior without weight updates to handle distribution shifts. \\
Tree-of-Thoughts Framework & ToT & A framework that enables exploration and evaluation of multiple reasoning paths via lookahead and backtracking, yielding gains in tasks requiring strategic planning. \\
United Nations Educational, Scientific and Cultural Organization & UNESCO & UN agency promoting global ethical standards for AI development, emphasizing equity, inclusiveness, and sustainability. \\
Vision Language Models & VLMs & Models that integrate visual perception and linguistic understanding for multimodal tasks, enabling capabilities such as visual question answering and image captioning. \\
Vision Transformer & ViT & Transformer architecture adapted for image recognition by treating image patches as sequence tokens, achieving state-of-the-art performance. \\
Voyager & Voyager & Open-ended embodied agent using large language models for autonomous exploration and skill acquisition in minecraft environments. \\
\end{longtable}
\normalsize
%%%% Enter Prompts here %%%
% \clearpage
% 
% Your table code here (the definitions table at the top)

\begin{figure}[h]
    \centering
    \includegraphics[width=0.9\linewidth]{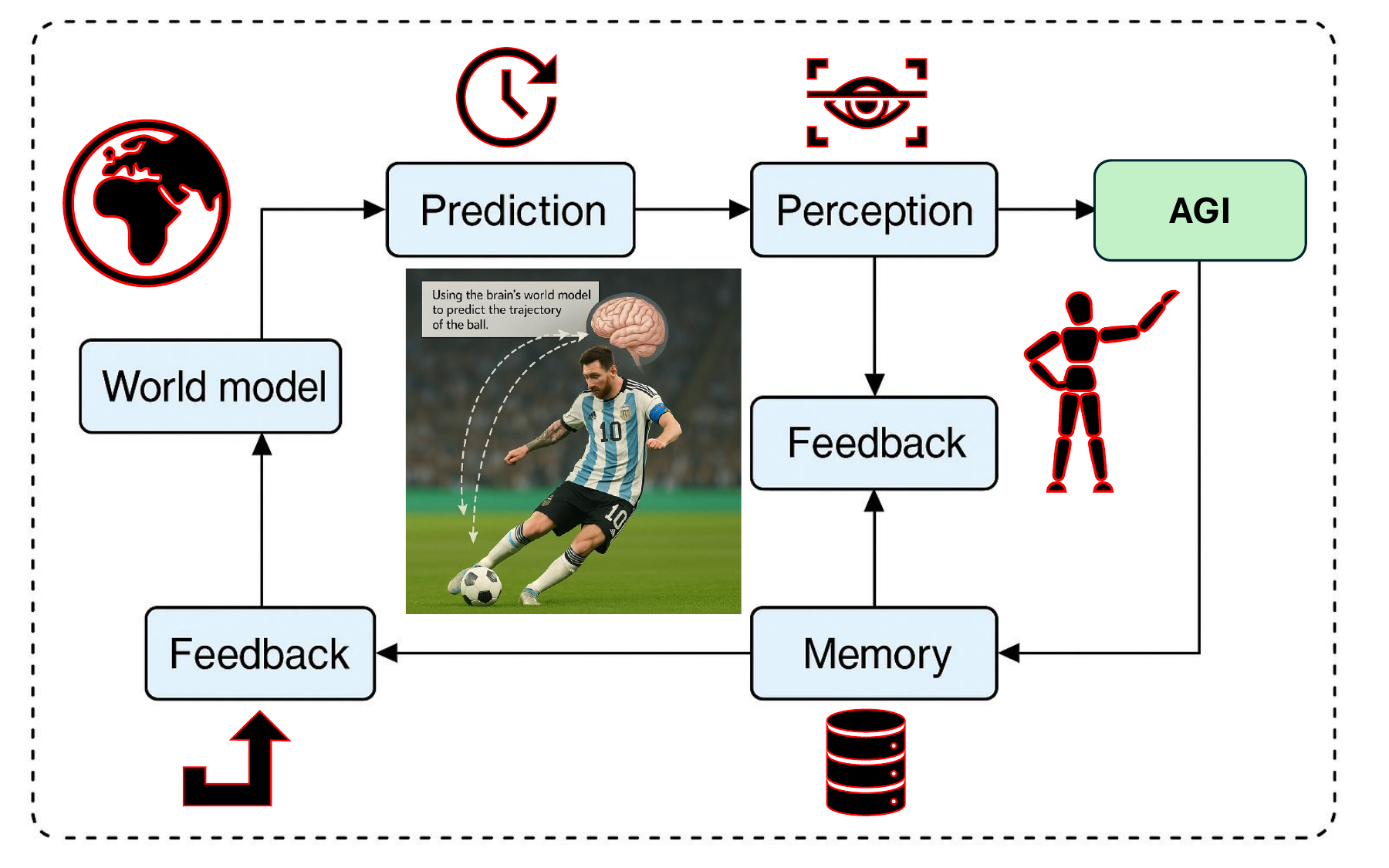}
    \caption{ A conceptual framework illustrating how world models, prediction, perception, memory, and feedback interact to enable intelligent behavior. The diagram reflects a biologically inspired loop where internal world models allow an agent to anticipate outcomes, update its beliefs through perception and memory, and adapt actions based on feedback. This architecture mirrors human cognition and serves as a foundational blueprint for developing AGI.}
    \label{fig:world_model_framework}
\end{figure}

\begin{figure}[ht!]
    \centering
    \begin{tcolorbox}[
        colback=yellow!10!white, 
        colframe=orange!70!black, 
        top=4pt, 
        bottom=4pt, 
        left=6pt, 
        right=6pt,
        title=\textbf{Task Description and Prompt Example},
        fonttitle=\bfseries\large,
        width=0.95\textwidth
    ]  
    \small  
    \textbf{Problem:} A company's sales increased from \$1.2M to \$1.5M over a quarter. The marketing budget in the same period went from \$200K to \$300K. What is the percentage increase in sales per dollar spent on marketing?
    \vspace{0.3cm}
    \textbf{Assistant's Response:}  
    First, compute the sales-to-marketing ratio before and after the increase:  
    \begin{align}
    \text{Initial ratio} &= \frac{1{,}200{,}000}{200{,}000} = 6 \\
    \text{Final ratio} &= \frac{1{,}500{,}000}{300{,}000} = 5
    \end{align}
    Now calculate the percentage change in efficiency (sales per dollar):  
    \[
    \frac{5 - 6}{6} \times 100 = -16.67\%
    \]  
    \textbf{Final Answer:} Sales efficiency decreased by 16.67\%. 
    \textit{Note: This example demonstrates the need for multimodal reasoning capabilities in AGI systems.}
    \end{tcolorbox}
    
    \caption{\textbf{ An example of multimodal reasoning in AGI systems}}
    \label{fig:supplementary_example}
\end{figure}

\begin{figure*}[ht!]
    \centering
    \includegraphics[width=0.75\textwidth]{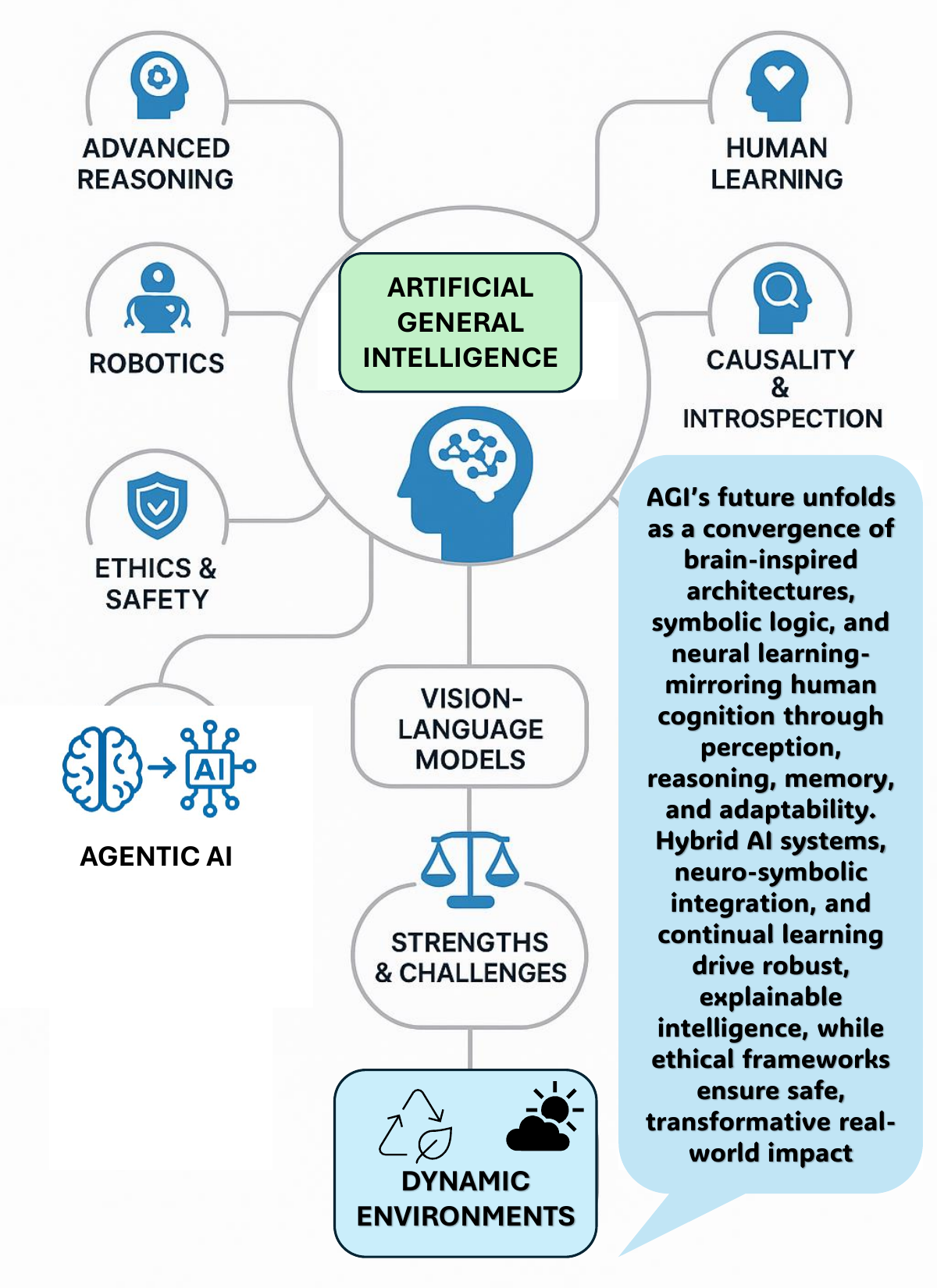}
    \caption{ AGI Development Roadmap: Illustrating a scientific roadmap of AGI development, highlighting hybrid AI architectures, core cognitive functions, memory systems, perception models, and ethical safeguards. The diagram shows how neuroscience and AI converge to shape generalizable, human-aligned artificial intelligence.}
    \label{fig:agi_roadmap}
\end{figure*}

\begin{figure*}[ht!]
    \centering
    \includegraphics[width=1.0\textwidth]{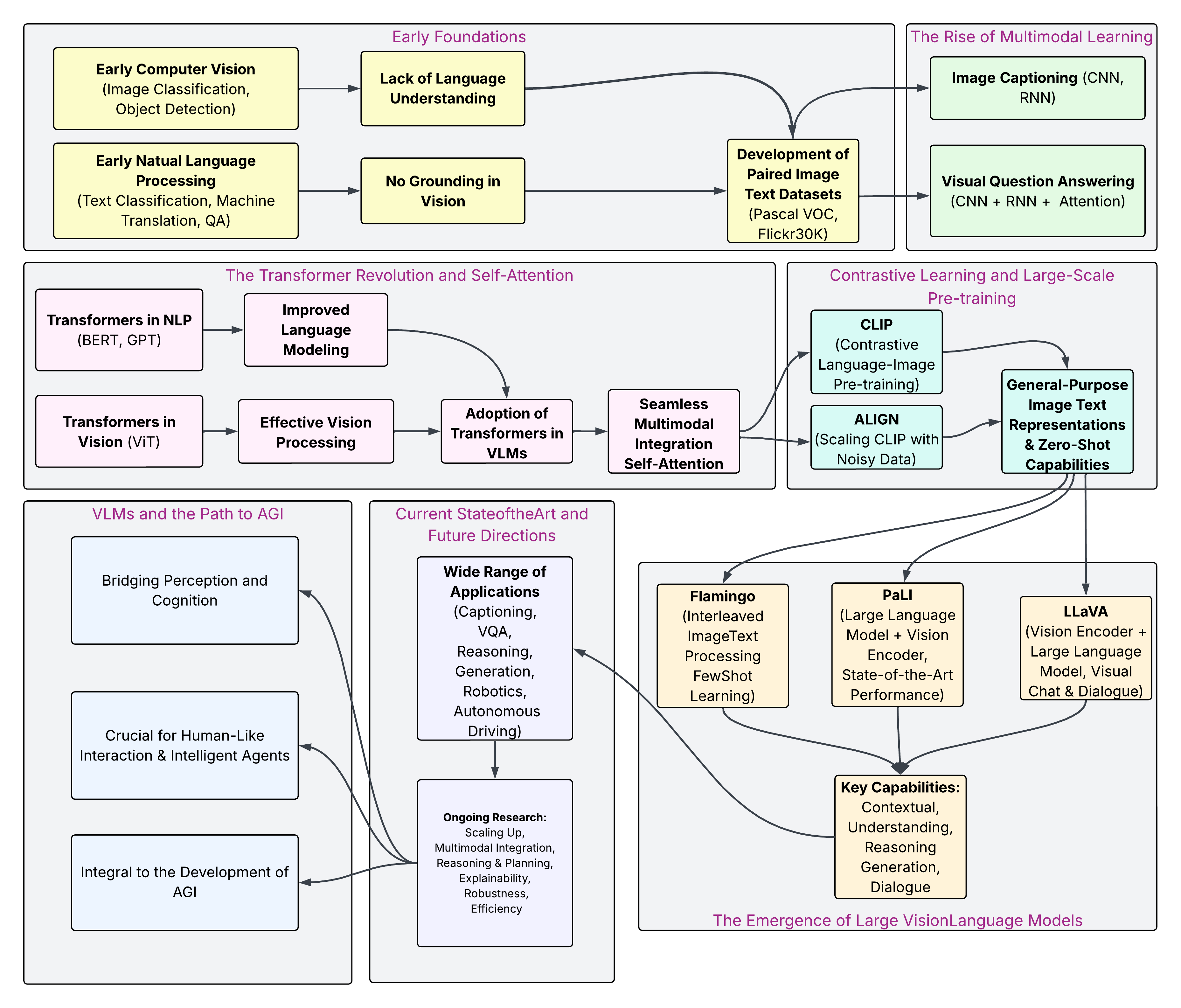}
    \caption{Conceptual roadmap tracing the evolution of VLMs. The figure outlines the progression from early unimodal systems in computer vision and natural language processing to modern VLMs enabled by self-attention, contrastive learning, and large-scale pretraining. It highlights pivotal developments such as paired image-text datasets, the adoption of transformers, and the emergence of general-purpose models like CLIP and ALIGN. The diagram also emphasizes the capabilities, applications, and future research directions of VLMs, positioning them as foundational components in the pursuit of AGI.}
    \label{fig:vlm}
\end{figure*}

\begin{figure*}[ht!]
    \centering
    \includegraphics[width=1.0\textwidth]{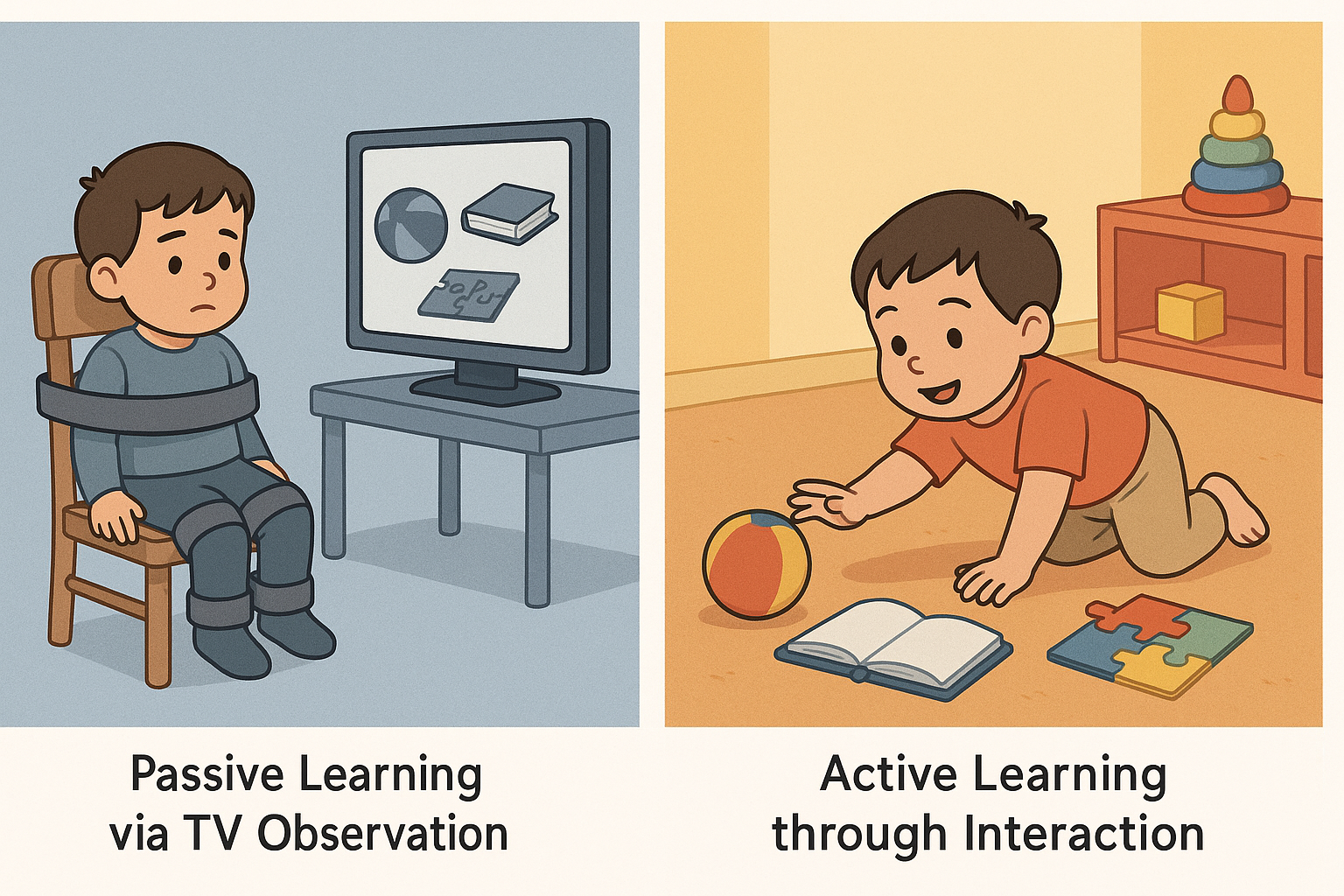}
    \caption{(a) Passive Learning via TV Observation. A child is depicted strapped to a chair, unable to move, with their attention focused on a television screen and b): Active Learning through Interaction. The same child is now free, crawling on the floor, and actively interacting with physical versions of the objects shown on the TV in (a).}
    \label{fig:child}
\end{figure*}

\end{document}